\documentclass[11pt,a4paper]{article}

\usepackage[T1]{fontenc}
\usepackage[utf8]{inputenc}
\usepackage{mathtools}
\usepackage{amssymb}
\usepackage{amsthm}
\usepackage{newtxtext,newtxmath}
\usepackage[a4paper,margin=1in]{geometry}
\usepackage{microtype}

\usepackage{graphicx}
\usepackage[dvipsnames]{xcolor}
\usepackage{siunitx}
\usepackage{booktabs}
\usepackage{multirow}
\usepackage{array}
\usepackage{subfig}

\usepackage[numbers,sort&compress]{natbib}
\usepackage{url}
\usepackage{verbatim}
\usepackage{authblk}
\usepackage[colorlinks=true,
            linkcolor=MidnightBlue,
            citecolor=MidnightBlue,
            urlcolor=MidnightBlue]{hyperref}

\graphicspath{{./}{../}}

\newif\ifshowrevisions
\showrevisionsfalse
\ifshowrevisions
  \newcommand{\rev}[1]{{\color{blue}#1}}
  
\else
  \newcommand{\rev}[1]{#1}
  
\fi
\newcommand{\len}{\text{\texttt{len}}}
\newcommand{\inc}{\text{\texttt{inc}}}

\DeclarePairedDelimiter{\round}\lfloor\rceil
\DeclareMathOperator*{\argmin}{arg\,min}

\DeclarePairedDelimiter\clip{\texttt{clip}(}{)}

\newcommand{\tol}{\text{\texttt{tol}}}

\newtheorem{theorem}{Theorem}
\newtheorem{proposition}[theorem]{Proposition}
\theoremstyle{definition}

\theoremstyle{remark}
\newtheorem{remark}{Remark}

\providecommand{\sep}{\unskip,\space}
\newenvironment{keywords}{%
  \par\vspace{0.6em}\noindent\textbf{Keywords: }\ignorespaces
}{\par\vspace{0.8em}}

\title{\bfseries QABBA: Error-Guaranteed Symbolic Time Series Compression\\
via Integer-Quantized Aggregation}

\author[1]{Erin Carson}
\author[2]{Xinye Chen}
\author[3]{Fei He}
\author[4]{Cheng Kang}

\affil[1]{Department of Numerical Mathematics, Charles University, Prague, Czech Republic\\
\texttt{carson@karlin.mff.cuni.cz}}
\affil[2]{LIP6, Sorbonne Université, CNRS, Paris, France}
\affil[3]{Centre for Computational Science and Mathematical Modelling, Coventry University, Coventry, United Kingdom\\
\texttt{fei.he@coventry.ac.uk}}
\affil[4]{Department of Cybernetics, Czech Technical University in Prague, Prague, Czech Republic\\
\texttt{kangchen@fel.cvut.cz}}

\date{}

\hypersetup{
  pdftitle={QABBA: Symbolic Time Series Compression via Integer-Quantized Aggregation},
  pdfauthor={Erin Carson, Xinye Chen, Fei He, Cheng Kang}
}

\begin{document}
\maketitle

\begin{center}
\small\emph{Authors are listed in alphabetical order and contributed equally.}
\end{center}
\vspace{0.5em}

\begin{abstract}
The rapid expansion of time-series data collected by sensors and monitoring systems has made compact representations increasingly important. These representations need to keep the main structure of the signal while cutting storage, transmission and computation costs. One effective route is to convert long numerical series into short symbolic sequences. Adaptive Brownian Bridge-based Aggregation (ABBA) is a recent method that does this well: it preserves the overall shape of the series and works for tasks such as anomaly detection and forecasting.

Three practical questions are still open. First, can the storage needed for ABBA's parameters be reduced further? Second, can the method run more efficiently if low-precision arithmetic is used? Third, can the symbolic sequences produced by ABBA be understood directly by large language models? \rev{We address these points with Quantized ABBA (QABBA), a quantized version of ABBA. By quantizing the symbolic centers, QABBA shrinks the parameter footprint and allows integer arithmetic, yet reconstruction quality remains high.} We also prove several error bounds that account for the extra approximation introduced by quantization: a dimension-free bound on the excess error of each segment, a reconstruction-error bound in the time domain, a stability condition for the symbolic assignment step, and a simple rule for allocating bits between segment lengths and increments.

\rev{Because the resulting symbolic strings are short and discrete, they can be passed straight to a pretrained large language model (LLM) without any extra time-series embedding layer. Experiments on the Monash regression archive, the UCR Time Series Classification Archive, and the UEA Multivariate Time Series Classification Archive show that the method offers a usable compromise among storage size, reconstruction accuracy and downstream predictive performance. These results show that QABBA provides an effective low-precision symbolic representation for time-series signals, while also supporting LLM-based time-series analysis.}

\rev{Our software is made publicly accessible in \url{https://github.com/inEXASCALE/qabba}. }
\end{abstract}


\section{Introduction}

The ubiquity of temporally annotated data produced, for example, by modern sensing, monitoring, and Internet of Things (IoT) systems, demands efficient storage and transmission of time series data. Time series often contain large numbers of samples and complex temporal structures, making their storage, processing, and analysis computationally demanding. The curse of dimensionality~\cite{10.1145/276698.276876} limits almost all data mining and machine learning algorithms when  dealing with large-scale data since they scale poorly with dimensionality.  Therefore, to store and analyze vast amounts of data, efficient management and mining of time series are critical in modern science and engineering domains; see \cite{8012550} for a survey. Numerous data mining and compression techniques have been developed to aid time series analysis in an elegant manner; see \cite{10.1145/2379776.2379788} and \cite{10.1145/3560814} for reviews. Computing a representation that reduces the effective signal length while preserving its essential temporal characteristics can improve algorithm efficiency and substantially reduce computational and storage requirements. Well-established representations include functional approximation methods based on the discrete Fourier transform~\cite{1311194}, discrete wavelet transform~\cite{10.14778/1687627.1687721}, subband coding \cite{FRANCIS19991895}, and piecewise linear representations~\cite{10.1145/568518.568520}, as well as \emph{symbolic approximation} methods such as Symbolic Aggregate approXimation (SAX)~\cite{lin2003symbolic}, one-dimensional SAX (1d-SAX)~\cite{10.1007/978-3-642-41398-8_24}, Adaptive Brownian Bridge-based symbolic Aggregation (ABBA)~\cite{EG19b} and fast ABBA (fABBA)~\cite{fABBA2022}.

As discussed in \cite{lin2007experiencing},  numerical representations are real-valued, which can limit the potential algorithms and data structures that can be applied directly. An alternative is to use symbolic time-series representation, whose efficacy has been demonstrated in a wide range of downstream tasks, including anomaly detection \cite{EG19b, conf/edbt/Senin0WOGBCF15}, classification \cite{10.1007/978-3-031-24378-3_4}, forecasting \cite{EG20b, CRIADORAMON2022108871}, motif discovery \cite{10.1145/1814245.1814255, Gao2019, 10.1145/956750.956808}, suffix tree mapping \cite{lin2004visually}, regular expression matching \cite{10.2312:eurova.20221081}, event prediction \cite{7953302}, and clustering \cite{8280846}. In this work, we focus on the ABBA \cite{EG19b}, a recently developed symbolic approximation technique. ABBA combines adaptive piecewise-linear compression with clustering-based digitization to reduce the effective length of a time series and convert it into a compact symbolic sequence. This representation preserves important temporal shape information and enables subsequent analysis in a lower-dimensional symbolic space. \rev{However, the symbolic centers required for reconstruction are still stored in floating-point format, which limits the potential gains in storage efficiency and low-precision computation.}


The rapid growth of large-scale time-series datasets, including those containing more than 100 million data samples in applications such as IoT sensing \cite{8012550} and astronomy \cite{6853469}, creates substantial challenges for efficient data storage, processing, and analysis. Although modern data-management systems can accommodate large volumes, processing such signals in real time or on resource-constrained devices remains difficult because of limitations in memory, bandwidth, and computational capacity. \rev{To address these challenges, we propose Quantized ABBA (QABBA), which converts time series into a compact symbolic representation while quantizing the associated symbolic centers to low-bitwidth integers. QABBA is designed to achieve a favorable balance between compression efficiency and reconstruction fidelity, thereby supporting scalable time-series analysis and low-precision implementation.} \rev{Using 8-bit representations for segment lengths, 12-bit representations for increments, and byte-sized symbols, QABBA can reduce the storage required for symbolic-center parameters by approximately two to ten times, depending on the representation setting.} This high compression rate is also consistent with Minimum Description Length (MDL) principles \cite{grunwald2007minimum}, as it aims to retain dominant temporal patterns (e.g., trends, anomalies) while discarding redundant or noise-like variations. \rev{Its utility is further evaluated through reconstruction experiments and LLM-based time-series regression tasks in Sections~\ref{sec:llm} and~\ref{sec:uea}.}

Our method is motivated by low-precision \emph{quantization} techniques widely used in modern signal-processing and machine learning systems. Quantization converts floating point numbers into lower-bitwidth integer representations, thereby reducing storage and computational costs. Integer-based arithmetic, such as integer matrix multiplication, can often be executed more efficiently than floating-point arithmetic on general-purpose processors, including x86 CPUs and ARM NEON architectures, and is particularly well supported by specialized hardware implementations such as UNPU \cite{8310262}, Eyeriss \cite{7551407}, Pragmatic \cite{10.1145/3123939.3123982} and NVIDIA Tensor cores. Related surveys on neural network quantization can be found in \cite{10.1145/3623402, 10028742}. Owing to its simplicity, low-bitwidth quantization can be integrated directly into ABBA-based methods without changing their main compression and digitization procedures or substantially compromising their downstream applications.
Moreover, if ABBA generates a large number of symbols, the reconstructed time series may be more accurate, but the symbolic representation is also more likely to encode noise and minor fluctuations in the original signal. This increases the representation complexity and computational cost of downstream models (e.g., LLMs), which is consistent with the MDL principle. By introducing quantization, QABBA reduces the storage required for the symbolic centers and enables integer-based operations in reconstruction and downstream applications that use the center information, such as time-series forecasting. Integer operations are generally supported by simpler computational pathways than floating-point operations on many general-purpose processors \cite{10.5555/3207796}, and may therefore offer improved computational efficiency on appropriate hardware. Furthermore, the reduced storage requirements of the QABBA parameters decrease the volume of data that must be transferred to downstream models, potentially reducing the training and inference costs of large language models.

\rev{The discrete symbolic sequence enables efficient string-based algorithms, such as motif discovery and anomaly detection, while the quantized centers support integer-based computation. On suitable low-power hardware, integer arithmetic can reduce processing time by approximately 2--3$\times$ compared with floating-point alternatives.}
Beyond storage savings, QABBA’s representation is particularly suitable for real-time processing in distributed and edge-computing environments, where bandwidth and memory are limited. For example, transmitting QABBA’s compact representation over low-bandwidth networks (e.g., Narrowband IoT) or performing on-device analytics on IoT sensors may be more practical than transmitting or processing the complete raw signal or a less compact representation. These advantages make QABBA a potentially powerful tool for applications such as anomaly detection in IoT sensor networks, classification and pattern analysis of financial and other large-scale time-series data, where preserving meaningful temporal structure while maintaining computational and storage efficiency is critical.

In this paper, we aim to develop a method that can approximate time-series signals using a meaningful and storage-efficient representation that can be applied to various downstream applications. To achieve this, we incorporate low-bitwidth quantization into ABBA-based symbolic approximation methods. The resulting framework, termed QABBA, preserves the main compression and digitization stages of ABBA while reducing the storage required for the symbolic centers and providing theoretical control over the additional error introduced by quantization. The primary contributions of this work are as follows:

{
\begin{enumerate}
    \item We introduce a low-bitwidth quantization strategy into the ABBA framework, resulting in a new method termed QABBA. By quantizing the symbolic centers while retaining the original compression and digitization stages, QABBA substantially reduces parameter storage with only a minor loss in approximation accuracy and negligible additional computational overhead. 
    \item We analyze the additional approximation error arising from quantization by relating the resulting sum of squared errors to the quantization error. In particular, we prove a dimension-free per-segment excess-\texttt{SSE} bound and an explicit two-dimensional bound for the length–increment representation used by QABBA. We further derive a bound on the propagation of increment quantization error to the reconstructed time-domain signal, a nearest-center stability condition for symbolic assignments, and a bit-allocation rule for selecting the length and increment precisions under a fixed storage budget. These results also apply more broadly to post-training center quantization in vector quantization and \texttt{k-means}.

    \item We investigate the use of QABBA with LLMs for time-series regression, demonstrating how compact symbolic pattern sequences can provide a discrete interface between numerical time-series signals and pretrained language models without requiring time-series embeddings to be learned from scratch.
    \item \rev{We empirically evaluate QABBA against established symbolic approximation methods, using several widely used datasets, in terms of quantization error across different bit-width settings, reconstruction quality, storage efficiency, and downstream regression performance.}
\end{enumerate}
}

The organization of the paper is as follows. Section~\ref{sec:related} reviews symbolic time-series representation methods and their applications. Section~\ref{sec:method} introduces the QABBA methodology, including the ABBA formulation, low-bitwidth quantization, theoretical error analysis, and symbolic reconstruction procedure. Section~\ref{sec:store} analyzes the storage requirements and compression efficiency
of ABBA, fABBA, and QABBA.  Section~\ref{sec:exp} presents extensive experimental evaluations on synthetic and real-world time-series datasets. Finally, Section~\ref{sec:final} summarizes the main findings and discusses future research directions.

\section{Related work}\label{sec:related}
This section will briefly review some well-established symbolic time-series representation (STSR) methods with dimensionality reduction, together with their representative applications. Given the extensive literature on STSR methods, we focus on a selection of widely used approaches that are most closely related to the proposed QABBA framework.

\rev{Symbolic Aggregate approXimation (SAX) \cite{lin2003symbolic} laid the foundation for using low-dimensional symbolic representations in numerous downstream time-series mining tasks, since it enables efficient indexing with a lower-bounding distance measure. SAX first applies Piecewise Aggregate Approximation (PAA) \cite{10.1145/568518.568520} to divide a continuous time series into equal-length segments; the segment averages form a reduced-dimensional PAA representation.} SAX then converts the PAA representation into a symbolic representation.  SAX-related time series mining methods have achieved significant success, e.g., in clustering (SAX Navigator \cite{8933618}, pattern search (SAXRegEx \cite{YU202313}), SPF \cite{10.1007/s10618-021-00798-w}), anomaly detection (HOT SAX \cite{1565683}, TARZAN \cite{10.1145/775047.775128}) and time series classification (SAX-VSM \cite{6729617}, BOPF \cite{8215500},  MrSQM \cite{10.1007/978-3-031-24378-3_4}). Numerous variants of SAX have been proposed to improve performance, e.g., 1d-SAX \cite{10.1007/978-3-642-41398-8_24}, pSAX \cite{9774017}, cSAX \cite{9774017}. Their novelty is mainly focused on reconstruction accuracy, but SAX is still a dominant symbolic representation method due to its appealing simplicity and speed.

\rev{Compared with SAX, ABBA symbolic approximation \cite{EG19b} achieves better reconstruction and pattern recognition for time series.} It relies on adaptive polygonal chain approximation followed by clustering to achieve the symbolization of time series. The reconstruction error of the representation can be modeled as a random walk with pinned start and end points, i.e., a \emph{Brownian bridge}. \rev{fABBA \cite{fABBA2022}, a variant of ABBA, uses a computationally efficient greedy aggregation (GA) method to speed up digitization by an order of magnitude. Both ABBA and fABBA have been empirically demonstrated to preserve the shape of time series better than SAX, especially the up-and-down behavior of time series. The application of ABBA has been shown to be effective for time-series prediction and anomaly detection; e.g., LSTM models with ABBA show robust inference performance \cite{EG20b}, LLM-ABBA \cite{11397443} framework demonstrates ABBA applications in  time series classification / forecasting / regression tasks,  and TARZAN with SAX replaced by ABBA or fABBA compares favorably with SAX-based TARZAN \cite{EG19b, fABBA2022}.} A downside of ABBA and fABBA is that they are unable to convert multiple time series in a symbolically unified manner. Recent work \cite{chen2024joint} addresses this issue by introducing a joint representation that enables a consistent symbolization.

\section{Methodology}\label{sec:method}

In this section, we will briefly recap the ABBA method and establish some notation for the QABBA method which we will use throughout the rest of the paper. As mentioned before, ABBA is an efficient STSR method based on an adaptive polygonal chain approximation followed by a means-based clustering algorithm. ABBA symbolization contains two main steps, namely \emph{compression} and \emph{digitization}, to aggregate a time series $T = [ t_1, t_2, \ldots, t_n] \in \mathbb{R}^{n}$ into a symbolic representation $A = [ a_1, a_2, \ldots, a_N ] \in \mathcal{L}^{N}$ where $\mathcal{L}$ denotes a finite alphabet set. A meaningful symbolic representation often has $N \ll n$.  The reconstruction error is measured by the distance between the original time series $T$ and the reconstructed time series $\widehat{T}$. It is evident that a bad symbolic approximation often leads to a high reconstruction error.

Our approach, QABBA, shares a similar procedure with ABBA, but with a quantization step inside the transform, which enables reduced storage space for time series representation.
The complete summary of notation for the QABBA procedure is listed in  \tablename~\ref{table:ABBA}. The procedures of \emph{symbolization} (the first three steps) and \emph{inverse-symbolization} (the last three steps) \footnote{For naming convenience, we define $\widehat{\inc}=\widetilde{\inc}$, following \cite{EG19b}.} are shared among ABBA and QABBA. QABBA contains two additional steps, quantization and inverse-quantization, which can potentially add to the reconstruction error. Note that in the table we use the term ``rounding'' for the seventh step instead of ``quantization'' as was used in \cite{EG19b} to avoid confusion.

\begin{table}[ht]
\caption{Summary of the QABBA procedure} 
\label{table:ABBA}
\centering \setlength\tabcolsep{1.7pt}
\begin{tabular}{l l}
\hline\\[-1mm]
time series & $T=[t_{0}, t_{1}, \ldots, t_{n}] \in \mathbb{R}^{n}$ \\[1mm]
after compression & $P=[(\len_{1}, \inc_{1}), \ldots, (\len_{N}, \inc_{N})] \in \mathbb{R}^{2 \times N}$ \\[1mm]
after digitization & $A=[a_{1}, \ldots, a_{N}] \in \mathcal{L}^{N}$, $C=[c_1, \ldots, c_k] \in \mathbb{R}^{2 \times k}$ \\[1mm]
after quantization & $A=[a_{1}, \ldots, a_{N}] \in \mathcal{L}^{N}$, $\widetilde{C}=[\tilde{c}_1, \ldots, \tilde{c}_k] \in \mathbb{N}^{2 \times k}$ \\[1mm]
inverse-quantization & $A=[a_{1}, \ldots, a_{N}] \in \mathcal{L}^{N}$,  $\widehat{C}=[\hat{c}_1, \ldots, \hat{c}_k] \in \mathbb{R}^{2 \times k}$ \\[1mm]
inverse-digitization & $\widetilde{P}=[(\widetilde{\len}_{1}, \widetilde{\inc}_{1}), \ldots, (\widetilde{\len}_{N}, \widetilde{\inc}_{N})] \in \mathbb{R}^{2 \times N}$ \\[1mm]
rounding & $\widehat{P}=[(\widehat{\len}_{1}, \widehat{\inc}_{1}), \ldots, (\widehat{\len}_{N}, \widehat{\inc}_{N})] \in \mathbb{R}^{2 \times N}$ \\[1mm]
inverse-compression & $\widehat{T}=[\widehat{t}_{1}, \widehat{t}_{2}, \ldots, \widehat{t}_{n}] \in \mathbb{R}^{n}$ \\ [1ex]
\hline 
\end{tabular}
\end{table}

\subsection{Compression}
In ABBA, compression is done by computing an adaptive piecewise linear continuous approximation of $T$ to obtain time series pieces $P = [(\len_{1}, \inc_{1}), \ldots, (\len_{N}, \inc_{N})] \in \mathbb{R}^{ N \times 2}$. The ABBA compression plays a central role in dimensionality reduction in ABBA symbolic approximation{---}a user-specific tolerance is set to control the degree of the reduction.

The ABBA compression proceeds by adaptively selecting $N+1$ indices $i_0 = 0 < i_1 <\cdots < i_N = n$ given a tolerance $\texttt{tol}$ such that the time series $T$ is well approximated by a polygonal chain going through the points $(i_j, t_{i_j})$ for $j=0,1,\ldots, N$. This results in a partition of $T$ into $N$ pieces $p_j=(\len_{j}, \inc_{j})$ that is determined by $T_{i_{j-1}:i_j} = [ t_{i_{j-1}},t_{i_{j-1}+1},\ldots, t_{i_j} ]$, each of integer length $\texttt{len}_j := i_j - i_{j-1}\geq 1$ in the time direction. \rev{Visually, each piece $p_j$ is represented by a straight line connecting the endpoint values $t_{i_{j-1}}$ and $t_{i_j}$. The partitioning criterion requires the squared Euclidean distance between the values in $p_j$ and the corresponding straight polygonal line to be upper bounded by $(\texttt{len}_j - 1)\cdot\texttt{tol}^2$.} For simplicity, given an index $i_{j-1}$ and starting with $i_0 = 0$, the procedure seeks the largest possible $i_j$ such that $i_{j-1} < i_j\leq n$ and
\begin{equation}\label{eq:compress}
\begin{aligned}
    \sum_{i=i_{j-1}}^{i_j} \Big( \, t_{i_{j-1}} + (t_{i_j} &- t_{i_{j-1}})\cdot \frac{i - i_{j-1}}{i_j - i_{j-1}}  - t_i \Big)^2 \\&\leq (i_{j} - i_{j-1} -1)\cdot\texttt{tol}^2.
\end{aligned}
\end{equation}

Each linear piece $p_j$ of the resulting polygonal chain is a tuple $(\texttt{len}_j, \texttt{inc}_j)$, where $\texttt{inc}_j = t_{i_j} - t_{i_{j-1}}$ is the increment in value, i.e., the subtraction of ending and starting value of $T_{i_{j-1}:i_j}$. The whole polygonal chain can be recovered exactly from the first value $t_0$ and the tuple sequence $p_1, p_2, \ldots, p_N$ where \rev{$p_i = (\texttt{len}_i, \texttt{inc}_i) \in \mathbb{R}^2$}. \rev{Because the reconstruction has pinned start and end points on each segment, its error can be naturally modeled as a Brownian bridge.}
In terms of \eqref{eq:compress}, a lower compression tolerance  value $\texttt{tol}$ is required to ensure a reasonable recovery error for the compression of time series with a great variety of features, e.g., trends, seasonal and nonseasonal cycles, pulses, and steps.

\subsection{Digitization}\label{subsec:digit}
The compression is followed by a digitization that leads to a \emph{symbolic representation} $A = [ a_1, a_2, \ldots, a_N ]\in \mathcal{L}^N$, $N\ll n$, and each $a_j$ is an element of a finite alphabet set $\mathcal{L}$ where $|\mathcal{L}| \ll N$. $\mathcal{L}$ can be referred to as a dictionary in the ABBA procedure.

Prior to digitizing, a normalization step is applied as indicated \cite{EG19b}; the lengths and increments are separately normalized by the factors $\sigma_{\texttt{len}}$ and $\sigma_{\texttt{inc}}$, respectively. \rev{The factors $\sigma_{\texttt{len}}$ and $\sigma_{\texttt{inc}}$ can be standard deviations or scaling factors chosen to map \texttt{len} and \texttt{inc} to a specific range such as $[-1, 1]$. Following normalization, a parameter $\texttt{scl}$ is employed to weight the length of each piece $p_i$ relative to its increment.} Hence, the clustering is effectively computed on
\begin{equation}\label{eq:tupseq2}
\begin{aligned}
\left(\texttt{scl}\frac{\texttt{len}_1}{\sigma_{\texttt{len}}}, \frac{\texttt{inc}_1}{\sigma_{\texttt{inc}}}\right), &\left(\texttt{scl}\frac{\texttt{len}_2}{\sigma_{\texttt{len}}}, \frac{\texttt{inc}_2}{\sigma_{\texttt{inc}}} \right), \hspace{2pt}
\ldots,\hspace{2pt} \left(\texttt{scl}\frac{\texttt{len}_N}{\sigma_{\texttt{len}}}, \frac{\texttt{inc}_N}{\sigma_{\texttt{inc}}} \right).
\end{aligned}
\end{equation}

In particular, if $\texttt{scl} = 0$, then clustering will be only performed on the increment values of $P$, while if $\texttt{scl} = 1$,  the lengths and increments are clustered with equal importance.

\rev{The steps after normalization work with vector quantization (VQ); VQ serves as a lossy data-compression technique that projects data into a low-dimensional space and plays an important role in signal processing.} Classical VQ is often achieved by means-based clustering. For ABBA symbolic approximation, Euclidean space is often assumed, which results in the Euclidean clustering problem. The notions of vector quantization are further explained in, e.g., \cite{1162229, 5075899}. Given an input of $N$ vectors $P=[p_1, \ldots, p_N] \in \mathrm{R}^{\ell\times N}$ where $\ell$ denotes dimensionality (in our context, $\ell = 2$), VQ seeks a codebook of $k$ vectors, i.e., $C=[c_1, \ldots, c_k] \in \mathrm{R}^{\ell \times k}$ such that $k$ is much smaller than $N$, where each $c_i$ is associated with a unique cluster $S_i$. \rev{A high-quality codebook keeps the sum of squared errors ($\texttt{SSE}$) close to its optimal value.} Suppose $k$ clusters $S_1, S_2, \ldots, S_k \subseteq P$ are computed. VQ aims to minimize
\begin{equation}\label{eq:sse}
    \texttt{SSE} = \sum_{i=1}^{k} \phi (c_i, S_i) = \sum_{i=1}^{k}\sum_{p\in S_i}\|p - c_i\|_2^2,
\end{equation}
where $\phi$ denotes the energy function and $c_i$ denotes the center of cluster $S_i$. We often compute the mean center $c_i = \frac{1}{|S_i|}\sum_{p \in S_i} p$ for the convergence of the energy function. The k-means algorithm \cite{journals/tit/Lloyd82} is a suboptimal solution to minimize $\texttt{SSE}$, i.e., to find $k$ cluster centers within the data in $\ell$-dimensional space to minimize \eqref{eq:sse}.  However, solving this problem exactly is NP-hard even if $k$ is restricted to $2$ \cite{10.1023/B:MACH.0000033113.59016.96} or in the plane \cite{MAHAJAN201213}. An alternative solution, used by the fABBA variant, is to use greedy aggregation (GA) \cite{fABBA2022} which gives an upper bound of $\alpha^2(N - k)$ for the $\texttt{SSE}$, where $\alpha$ is a hyper-parameter as described in \cite{fABBA2022}.

The ABBA digitization can be performed by a suitable partitional clustering algorithm that finds $k$ clusters from $P \in \mathbb{R}^{2 \times N}$ such that the $\texttt{SSE}$ given by $C$ is minimized. The obtained codebook vectors are known as cluster centers. In the context of symbolic approximation, we refer to the cluster centers as \emph{symbolic centers}. Each symbolic center is associated with an identical symbol and each time series snippet $p_i$ is assigned with the closest symbolic center $c^i$ associated with its symbol
\begin{equation}
    c^i = \argmin_{c \in C}(\|p - c\|).
\end{equation}

Correspondingly each symbolic center is projected to a symbol, written as $I_d: C \to A$. More formally, we define the digitization mapping $\psi: P \to A$ as
\begin{equation}\label{eq:digit}
    \psi(p_i) = I_d(c^i) = I_d(\argmin_{c \in C}(\|p - c\|)).
\end{equation}

\rev{Therefore, each symbol can be viewed as being associated with a unique cluster membership. In applications, each clustering label can be represented by a byte-sized integer value. The symbols used in ABBA can be represented by text characters and are not limited to English alphabet letters, since more clusters may be used. Each character in most computer systems can be associated with a byte-sized integer value, and symbolic strings can also be stored using other symbol encodings. To speed up digitization, existing work \cite{fABBA2022} uses GA to replace VQ, which results in a minor loss of approximation accuracy.} As analyzed in \cite{fABBA2022}, not all clustering approaches (see e.g., spectral clustering \cite{10.5555/946247.946658}, DBSCAN \cite{10.5555/3001460.3001507} and HDBSCAN \cite{10.1007/978-3-642-37456-2_14}) are suitable for the clustering in the digitization phase. \rev{This is particularly the case for density-based clustering methods, which often provide insufficient symbolic information to fully capture time-series patterns, since they can suffer from the chaining effect and are less likely to produce a satisfactory $\texttt{SSE}$, thus leading to high reconstruction error. In this paper, the performance of QABBA with VQ and GA is investigated.} However, some parallel implementations might bring potential speed improvement, e.g., parallel k-means \cite{FRANTI201995}. In addition, some methods like CNAK \cite{SAHA2021107625} may assist the selection of the $k$ parameter in ABBA by providing prior knowledge on the number of clusters (distinct symbols).

\subsection{Quantization and inverse-quantization}\label{sec:quanterr}

In the following, we formulate the quantization and inverse-quantization steps and then quantify the additional \texttt{SSE} caused by quantizing symbolic centers. The analysis is stated first for general vector quantization and then specialized to the two-dimensional \texttt{len}--\texttt{inc} arrays used by QABBA. This separation is useful because the same argument applies to post-training center quantization in \texttt{k-means} and other means-based clustering methods.

The scalar \emph{quantization} used here is governed by a scale factor $s \in \mathbb{R}^{+}$, a zero-point $z$, and a bit-width $\omega$. A quantization mapping $Q: \mathbb{R} \to \mathbb{N}$ maps a floating point number $x \in [\zeta, \eta]$ to a $\omega$-bit integer $\tilde{x} \in [\widetilde{\zeta}, \widetilde{\eta}]$. The bit-width $\omega$ determines the integer range; for signed integer quantization we use $\widetilde{\zeta} = -2^{\omega - 1}$ and $\widetilde{\eta} = 2^{\omega - 1} - 1$. The quantization mapping is given by
\begin{equation}\label{eq:quant1}
    \tilde{x} = Q(x) = \round{\frac{1}{s} x - z},
\end{equation}
where $s = \frac{\eta - \zeta}{\widetilde{\eta} - \widetilde{\zeta}}=\frac{\eta-\zeta}{2^\omega-1}$, and $\round{\cdot}$ is the operator that rounds the value to the nearest integer.

Often, clipping is employed for the quantized value, that is,
\begin{equation*}
    \clip{x, a, b} = \begin{cases}
        a & \text{if } x < a\\
        x & \text{if } a < x < b\\
        b & \text{if } x > b
    \end{cases}.
\end{equation*}

Hence \eqref{eq:quant1} becomes
\begin{equation}\label{eq:quant2}
    \tilde{x} =  Q(x) = \clip{\round{\frac{1}{s} x - z}, a, b}.
\end{equation}

If $a, b$ are exactly $\widetilde{\zeta}, \widetilde{\eta}$, respectively, and all symbolic centers lie in the calibrated interval $[\zeta,\eta]$, then the clipping operation is inactive. Accordingly, the \emph{inverse-quantization} is defined as
\begin{equation}\label{eq:dequant}
    y = Q^{-1}(\tilde{x}) = s(\tilde{x} + z)  = s(\clip{\round{\frac{1}{s} x - z}, a, b} + z).
\end{equation}

When the zero-point $z$ is set to zero, one obtains symmetric quantization, a case of particular interest because it avoids the overhead of handling the zero-point offset \cite{Jacob_2018_CVPR}. In what follows we assume that floating point roundoff is negligible compared with the quantization step; this is valid when the working precision unit roundoff is much smaller than $s$. If clipping is inactive, nearest-integer rounding gives
\begin{equation}
    \round{f}=f+\rho,\qquad |\rho|\leq \frac{1}{2}.
\end{equation}

Combining this model with \eqref{eq:dequant}, the dequantized value $y=Q^{-1}(Q(x))$ satisfies
\begin{equation}
    y=x+s\rho,\qquad |y-x|\leq \frac{s}{2}=\frac{\eta-\zeta}{2(2^\omega-1)}=\frac{\eta-\zeta}{2^{\omega+1}-2}.
\label{eq:scalar_quant_error}
\end{equation}

Accordingly, the symbolic centers $C=[c_1,\ldots,c_k]$ are quantized as $\widetilde{C}:=Q(C)$ and then inverse-quantized as $\widehat{C}:=Q^{-1}(\widetilde{C})$. It is $\widehat{C}$, not the integer-valued $\widetilde{C}$, that is used during symbolic reconstruction. The next result shows how the center quantization error propagates to the clustering \texttt{SSE}.

\begin{theorem}[Excess \texttt{SSE} after center quantization]\label{thm:quant_sse_general}
Let $P=\{p_1,\ldots,p_N\}\subset \mathbb{R}^{\ell}$ be partitioned into nonempty clusters $S_1,\ldots,S_k$, and let each center be the cluster mean $c_i=|S_i|^{-1}\sum_{p\in S_i}p$. Define
\begin{equation*}
    \texttt{SSE}=\sum_{i=1}^{k}\sum_{p\in S_i}\|p-c_i\|_2^2,\qquad
    \widehat{\texttt{SSE}}=\sum_{i=1}^{k}\sum_{p\in S_i}\|p-\widehat{c}_i\|_2^2,
\end{equation*}
where $\widehat{c}_i=Q^{-1}(Q(c_i))$ is the inverse-quantized center. Then
\begin{equation}\label{eq:sse_identity}
    \widehat{\texttt{SSE}}-\texttt{SSE}
    =\sum_{i=1}^{k}|S_i|\|\widehat{c}_i-c_i\|_2^2.
\end{equation}
Consequently, if $\|\widehat{c}_i-c_i\|_2\leq \varepsilon_i$ for all $i$, then
\begin{equation}\label{eq:sse_general_bound}
    \widehat{\texttt{SSE}}\leq \texttt{SSE}+\sum_{i=1}^{k}|S_i|\varepsilon_i^2.
\end{equation}
In particular, if a quantizer is chosen so that $\|\widehat{c}_i-c_i\|_2\leq \varepsilon$ for every center, then
\begin{equation}\label{eq:dimension_free_bound}
    \frac{\widehat{\texttt{SSE}}-\texttt{SSE}}{N}\leq \varepsilon^2.
\end{equation}
The per-segment excess-\texttt{SSE} bound \eqref{eq:dimension_free_bound} is independent of the ambient dimension $\ell$, the number of segments $N$, and the number of centers $k$, except through the prescribed center-error tolerance $\varepsilon$.
\end{theorem}

\begin{proof}
For any $p\in S_i$, write $p-\widehat{c}_i=(p-c_i)+(c_i-\widehat{c}_i)$. Summing the squared norms over $S_i$ gives
\begin{equation*}
\begin{aligned}
\sum_{p\in S_i}\|p-\widehat{c}_i\|_2^2
=&\sum_{p\in S_i}\|p-c_i\|_2^2
+|S_i|\|c_i-\widehat{c}_i\|_2^2\\&
+2(c_i-\widehat{c}_i)^T\sum_{p\in S_i}(p-c_i).
\end{aligned}
\end{equation*}
The cross term is zero because $c_i$ is the mean of $S_i$. Summing over $i=1,\ldots,k$ proves \eqref{eq:sse_identity}; \eqref{eq:sse_general_bound} and \eqref{eq:dimension_free_bound} follow immediately.
\end{proof}

The theorem does not require the centers to be globally optimal. Hence it applies to ABBA with \texttt{k-means}, to fABBA with greedy aggregation, and more generally to any VQ procedure that stores a fixed assignment together with representative centers. If, after quantization, one reassigns each point to the nearest inverse-quantized center as in the standard \texttt{k-means} objective, the resulting objective can only be smaller than $\widehat{\texttt{SSE}}$ computed with the fixed symbolic assignments, so the same upper bound remains valid.

\begin{proposition}[Two-dimensional QABBA bound]\label{prop:qabba_2d_sse}
Consider QABBA centers $c_i=(c_{i,\len},c_{i,\inc})^T\in\mathbb{R}^2$. Suppose that the \texttt{len} and \texttt{inc} coordinates are quantized independently with bit-widths $\omega_{\len}$ and $\omega_{\inc}$ over calibrated ranges $[\zeta_{\len},\eta_{\len}]$ and $[\zeta_{\inc},\eta_{\inc}]$, with inactive clipping. Let
\begin{equation*}
    \Delta_{\len}=\frac{\eta_{\len}-\zeta_{\len}}{2^{\omega_{\len}}-1},
    \qquad
    \Delta_{\inc}=\frac{\eta_{\inc}-\zeta_{\inc}}{2^{\omega_{\inc}}-1}.
\end{equation*}
Then
\begin{equation}\label{eq:qabba_2d_bound}
    \widehat{\texttt{SSE}}\leq
    \texttt{SSE}
    +N\left(\frac{\Delta_{\len}^2}{4}+\frac{\Delta_{\inc}^2}{4}\right),
    \qquad
    \frac{\widehat{\texttt{SSE}}-\texttt{SSE}}{N}
    \leq
    \frac{\Delta_{\len}^2+\Delta_{\inc}^2}{4}.
\end{equation}
If min--max normalization maps both coordinates to $[-1,1]$, then
\begin{equation}\label{eq:qabba_2d_normalized_bound}
    \frac{\widehat{\texttt{SSE}}-\texttt{SSE}}{N}
    \leq
    \frac{1}{(2^{\omega_{\len}}-1)^2}
    +
    \frac{1}{(2^{\omega_{\inc}}-1)^2}.
\end{equation}
For the bit-widths used in our experiments, $\omega_{\len}=8$ and $\omega_{\inc}=12$, the normalized per-segment excess-\texttt{SSE} is at most $1.54\times 10^{-5}$.
\end{proposition}

\begin{proof}
By \eqref{eq:scalar_quant_error}, each inverse-quantized coordinate satisfies $|\widehat{c}_{i,\len}-c_{i,\len}|\leq \Delta_{\len}/2$ and $|\widehat{c}_{i,\inc}-c_{i,\inc}|\leq \Delta_{\inc}/2$. Therefore
\begin{equation*}
    \|\widehat{c}_i-c_i\|_2^2
    \leq \frac{\Delta_{\len}^2}{4}+\frac{\Delta_{\inc}^2}{4}
\end{equation*}
for every symbolic center. Applying Theorem~\ref{thm:quant_sse_general} gives \eqref{eq:qabba_2d_bound}. If both coordinates are normalized to $[-1,1]$, then $\eta_{\len}-\zeta_{\len}=\eta_{\inc}-\zeta_{\inc}=2$, which yields \eqref{eq:qabba_2d_normalized_bound}.
\end{proof}

Equations~\eqref{eq:dimension_free_bound} and \eqref{eq:qabba_2d_bound} clarify the role of $N$. The total \texttt{SSE} is an additive quantity and therefore naturally scales with the number of encoded segments, whereas the average increase per segment is controlled only by the quantization resolution. This is the relevant quantity when comparing datasets or symbolic representations with different lengths. The bounds also explain the empirical bit-width choice in Section~\ref{sec:exp}: the \texttt{len} coordinate tolerates fewer bits because its calibrated range is typically tighter and its reconstruction effect is moderated by the subsequent length-rounding step, while the \texttt{inc} coordinate benefits from a larger bit-width because it directly controls the vertical displacement of the reconstructed polygonal chain.

\begin{theorem}[Propagation to time-domain reconstruction error]\label{thm:time_domain_mse}
Consider a fixed symbolic representation with segment lengths $L_j\geq 1$ and increments $u_j$, $j=1,\ldots,N$, where $n=\sum_{j=1}^{N}L_j$. Assume that quantization changes only the increments, so that the quantized increments are $\widehat{u}_j=u_j+e_j$, while the segment lengths and the initial value are unchanged. Let $E_{j-1}=\sum_{m=1}^{j-1}e_m$ be the accumulated vertical displacement error before segment $j$. Then the additional squared error between the unquantized and quantized polygonal reconstructions satisfies
\begin{equation}\label{eq:time_domain_exact}
    \texttt{SSE}_{T}
    =
    \sum_{j=1}^{N}\sum_{r=1}^{L_j}
    \left(E_{j-1}+\frac{r}{L_j}e_j\right)^2 .
\end{equation}
Consequently, if $|e_j|\leq \varepsilon_{\inc}$ for all $j$, then
\begin{equation}\label{eq:time_domain_worst}
    \frac{\texttt{SSE}_{T}}{n}
    \leq
    2\varepsilon_{\inc}^2
    \left(
    1+\frac{1}{n}\sum_{j=1}^{N}L_j(j-1)^2
    \right)
    \leq
    2\varepsilon_{\inc}^2(1+N^2).
\end{equation}
Moreover, if $e_1,\ldots,e_N$ are independent, zero-mean errors with $\mathbb{E}e_j^2\leq\sigma_{\inc}^2$, then
\begin{equation}\label{eq:time_domain_expected}
    \frac{\mathbb{E}\texttt{SSE}_{T}}{n}
    \leq
    \sigma_{\inc}^2
    \left(
    1+\frac{1}{n}\sum_{j=1}^{N}L_j(j-1)
    \right)
    \leq
    N\sigma_{\inc}^2.
\end{equation}
\end{theorem}

\begin{proof}
Within segment $j$, at the relative location $r/L_j$, the unquantized polygonal reconstruction has accumulated all previous increments and the fraction $(r/L_j)u_j$ of the current increment. The quantized reconstruction differs by the accumulated previous increment error $E_{j-1}$ and the current within-segment error $(r/L_j)e_j$, which gives \eqref{eq:time_domain_exact}. For \eqref{eq:time_domain_worst}, use $(a+b)^2\leq 2a^2+2b^2$, $|E_{j-1}|\leq (j-1)\varepsilon_{\inc}$, and $r/L_j\leq 1$. For \eqref{eq:time_domain_expected}, independence and zero mean imply
\[
    \mathbb{E}\left(E_{j-1}+\frac{r}{L_j}e_j\right)^2
    =
    \mathbb{E}E_{j-1}^2+\frac{r^2}{L_j^2}\mathbb{E}e_j^2
    \leq
    (j-1)\sigma_{\inc}^2+\sigma_{\inc}^2.
\]
Summing over all segments and dividing by $n$ proves the result.
\end{proof}

\begin{remark}
The deterministic bound \eqref{eq:time_domain_worst} is deliberately conservative because it allows all increment errors to have the same sign and accumulate monotonically. The expected bound \eqref{eq:time_domain_expected} is often more descriptive of calibrated rounding quantization, where signs of rounding errors tend to cancel. In QABBA, $\varepsilon_{\inc}=\Delta_{\inc}/2$ under the inactive-clipping assumption of Proposition~\ref{prop:qabba_2d_sse}.
\end{remark}

\begin{proposition}[Stability of symbolic assignments]\label{prop:assignment_stability}
Let $C=\{c_1,\ldots,c_k\}$ be the symbolic centers and assume that each center is quantized with $\|\widehat{c}_i-c_i\|_2\leq\varepsilon$. For a point $p$, let $c_{a(p)}$ be its nearest unquantized center and let
\[
    \gamma(p)=\min_{i\neq a(p)}\|p-c_i\|_2-\|p-c_{a(p)}\|_2
\]
be its nearest-center margin. If $\gamma(p)>2\varepsilon$, then $p$ is assigned to the same symbolic center before and after center quantization.
\end{proposition}

\begin{proof}
For the original nearest center, $\|p-\widehat{c}_{a(p)}\|_2\leq \|p-c_{a(p)}\|_2+\varepsilon$. For any other center $c_i$, the reverse triangle inequality gives $\|p-\widehat{c}_{i}\|_2\geq \|p-c_i\|_2-\varepsilon$. If $\gamma(p)>2\varepsilon$, then $\|p-c_i\|_2-\varepsilon>\|p-c_{a(p)}\|_2+\varepsilon$ for all $i\neq a(p)$, so no competing quantized center is closer than $\widehat{c}_{a(p)}$.
\end{proof}

\begin{proposition}[Bit allocation under a fixed center budget]\label{prop:bit_allocation}
Suppose the \texttt{len} and \texttt{inc} coordinates are quantized independently and that a total of $B$ bits is available, so $\omega_{\len}+\omega_{\inc}=B$. Under the high-resolution approximation $2^\omega-1\approx 2^\omega$, minimizing the two-dimensional excess-\texttt{SSE} bound in \eqref{eq:qabba_2d_bound} is equivalent to minimizing
\[
    R_{\len}^2 2^{-2\omega_{\len}}+
    R_{\inc}^2 2^{-2\omega_{\inc}},
    R_{\len}=\eta_{\len}-\zeta_{\len},\quad
    R_{\inc}=\eta_{\inc}-\zeta_{\inc}.
\]
The continuous optimum satisfies
\begin{equation}\label{eq:bit_allocation}
    \omega_{\inc}-\omega_{\len}
    =
    \log_2\left(\frac{R_{\inc}}{R_{\len}}\right).
\end{equation}
Thus the coordinate with the larger calibrated range should receive more bits; in particular, when the increment range is larger or has greater reconstruction impact, allocating more bits to \texttt{inc} is theoretically justified.
\end{proposition}

\begin{proof}
Substitute $\omega_{\inc}=B-\omega_{\len}$ and minimize
\[
    f(\omega_{\len})=R_{\len}^2 2^{-2\omega_{\len}}+
    R_{\inc}^2 2^{-2(B-\omega_{\len})}.
\]
Setting $f'(\omega_{\len})=0$ gives
$R_{\len}^2 2^{-2\omega_{\len}}=R_{\inc}^2 2^{-2\omega_{\inc}}$.
Taking base-two logarithms yields \eqref{eq:bit_allocation}.
\end{proof}

\subsection{Symbolic reconstruction}\label{subsec:inverse_sb}
\rev{The \emph{symbolic reconstruction} (also called \emph{inverse symbolization}) refers to converting the symbolic representation $A$ into the reconstructed series $\widehat{T}$ so that the distance between $\widehat{T}$ and $T$, i.e., the error arising from symbolic approximation, is reasonably small.}
The symbolic reconstruction contains four dominant steps.
First, \rev{we inverse-quantize the integer-valued symbolic centers $\widetilde{C}$ to obtain $\widehat{C}$. Second, the \emph{inverse-digitization} is computed, simply written as $\psi^{-1}$ (see Equation~\eqref{eq:digit}), which uses the $k$ representative elements $\widehat{c}_i \in \widehat{C}$ (e.g., inverse-quantized mean or median centers of the groups $S$) to replace the symbols in $A$. This results in a 2-by-$N$ array $\widetilde{P}$, i.e., an approximation of $P$, where each $\widetilde{p}_i \in \widetilde{P}$ is associated with the symbolic center assigned to $p_i \in P$.}

\rev{The inverse digitization often leads to a non-integer value for the reconstructed length $\len$, so \cite{EG19b} proposes a rounding method to align the cumulated lengths with the closest integers, which is the third step. We round the first length to an integer value, i.e., $\widehat{\len}_1:= \text{round}(\widetilde{\len}_1)$, and calculate the rounding error $e:= \widetilde{\len}_1 - \widehat{\len}_1$. The error is then added to the rounding of $\widetilde{\len}_2$, i.e., $\widehat{\len}_2 := \text{round}(\widetilde{\len}_2 + e)$, and the new error $e'$ is calculated as $e'=\widetilde{\len}_2+e-\widehat{\len}_2$. Then $e'$ is used in the next rounding step in the same way.} After all rounding is computed, we obtain
\begin{equation}\label{eq:repolygon}
\widehat{P}=[(\widehat{\len}_{1}, \widehat{\inc}_{1}), \ldots, (\widehat{\len}_{N}, \widehat{\inc}_{N})] \in \mathbb{R}^{2 \times N},
\end{equation}
where the increments $\inc$ are unchanged, i.e., $\widehat{\inc} = \widetilde{\inc}$. The last step is to recover the whole polygonal chain exactly from the initial time value $t_0$ and the tuple sequence \eqref{eq:repolygon} via the inverse compression.

\rev{Table~\ref{table:ecg} provides a more intuitive understanding of the symbolic approximation achieved by QABBA. It shows three ECG (electrocardiogram) signals reconstructed using the QABBA variants based on vector quantization (VQ) and greedy aggregation (GA). The approximation is based on 8-bit quantization for segment length and 16-bit quantization for increment, with parameters set to $\texttt{tol} = 0.1$ and $\alpha = 0.4$. For each signal, the table presents the parameter settings, the resulting symbolic representation, and the number of distinct symbols (\#Symbols) used.} 

The inverse symbolization of ABBA unavoidably leads to some reconstruction errors due to the limited parameter space. Note that the reconstruction error is equivalent to the approximation error of the ABBA symbolization. The reconstruction error can be quantified via the mean squared error (MSE), which is given by
 \begin{equation}\label{eq:mse}
     \text{MSE} = \frac{1}{n}\sum_{i=1}^{n}  (t_i - \widehat{t}_{i})^2,
 \end{equation}
and its square root is referred to as root mean square error (RMSE).

\rev{Other measures, including Euclidean distance, Dynamic Time Warping (DTW) \cite{1163055}, and the differenced versions used in \cite{EG19b}, can also provide insights into the approximation error. For comprehensive evaluation, we use a combination of these measures in this paper.}

\begin{table}[ht]
\caption{Symbolic approximation of ECG signals.} 
\label{table:ecg} \scriptsize
\centering \setlength\tabcolsep{1.3pt}
\begin{tabular}{l l l l l}
\toprule\\[-1mm]
Signals & Parameter setting& Symbolic representation& \#Symbols & MSE \\[1mm]
\midrule
\multirow{ 4}{*}{I} & QABBA (VQ, scl=1) & AceDfCBdabEaGF & 13 & 0.12\\[1mm]
&QABBA (GA, scl=1) & ABCeDfGFEbdDca & 13 & 0.058\\[1mm]
&QABBA (VQ, scl=5) & AcFGDaECedbBfg & 14 & 0.058\\[1mm]
&QABBA (GA, scl=5) & baBcCGgfEeFdDA  & 14 & 0.058\\[1ex]
\midrule
\multirow{ 4}{*}{II} & QABBA (VQ, scl=1) & AbdCDBEacECbeFfA &12 & 0.32\\[1mm]
& QABBA (GA, scl=1) & AbeEFfdDadcbBCcA &12 & 0.073\\[1mm]
& QABBA (VQ, scl=5) & AHFECagdDbcBfeGA &15 & 0.069\\[1mm]
& QABBA (GA, scl=5) & ACEdgHGeFfcDbaBA &15 & 0.12\\[1ex]
\midrule
\multirow{ 4}{*}{III} & QABBA (VQ, scl=1) & ADCadcBeEbdC &10 & 0.29\\[1mm]
& QABBA (GA, scl=1) & AabdcEeCBDcb &10 & 0.12\\[1mm]
& QABBA (VQ, scl=5) & AEcdbBaeCDbc &10 & 0.069\\[1mm]
& QABBA (GA, scl=5) & aBAbCEecDdCA &10 & 0.12\\[1ex]
\bottomrule
\end{tabular}
\end{table}

\section{Storage reduction}\label{sec:store}
\rev{As discussed previously, the ABBA method and its variants are symbolic approximation techniques that achieve efficient data compression, significantly reducing the storage needs of time series. Reconstructing an ABBA representation requires the initial value(s), symbolic center(s), and strings (we refer to Section~\ref{sec:method} for more detail). Therefore, to store the time-series representation, only these three variables are needed. Moreover, when the number of distinct symbols in an STSR exceeds a data-dependent threshold, the storage of ABBA is string-dominant, where the number of string symbols is the primary factor; otherwise, it is symbolic-center-dominant. In the string-dominant case, further compression can be achieved using lossless coding of the symbolic sequence, for example, the Lempel-Ziv-Welch (LZW) algorithm \cite{1659158}.}

Time series are often represented as floating point numbers, primarily 32 bits, i.e., single-precision format (abbreviated as float32). To store the time series, we require $B_T$ bits for each value. \rev{In practice, ABBA uses a Unicode character corresponding to a given cluster center, i.e., a distinct symbol. Internally, following PEP 393, Python uses a flexible storage strategy that dynamically chooses the most efficient encoding: 1 byte for ASCII (Latin-1), 2 bytes for BMP characters (UTF-16), and 4 bytes for supplementary characters (UTF-32). While the character data itself may use 1--4 bytes depending on the code point, the overall memory usage of the resulting string object is higher due to metadata and object overhead. Because ABBA normally uses ASCII symbols, we assume for simplicity that one symbol takes one byte.} Let's assume $B_{\texttt{len}}$ and $B_{\texttt{inc}}$ bits are used to represent each \texttt{len} and \texttt{inc} value of the symbolic centers of length $k$, respectively. \rev{The dominant storage of an ABBA symbolic representation associated with $p$ initial values thus requires $8N + (B_{\texttt{len}} + B_{\texttt{inc}}) \cdot k + p\cdot B_T$ bits. We define the storage ratio of a time series of length $n$ as the ratio between the compressed size and the uncompressed size,}
\begin{equation}\label{eq:compress_rate1}
\rev{\varphi_{ABBA}  = \frac{8N + (B_{\texttt{len}} + B_{\texttt{inc}}) \cdot k + p\cdot B_T }{B_T \cdot n} .}
\end{equation}

For ABBA approximation, if float32 is used for the time series and all values of symbolic centers, then \eqref{eq:compress_rate1} can be computed as $\varphi_{ABBA}  = \frac{ 8N + 64 k + 32}{32 n}$, which also holds for the fABBA variant.

\rev{Compared to ABBA, QABBA requires additional storage for the quantization scale and zero-point parameters for the \texttt{len} and \texttt{inc} coordinates. If each of these four parameters uses $B_s$ bits, the additional storage is constant complexity and thus does not significantly affect the storage ratio. The QABBA storage ratio $\varphi_{QABBA}$ can similarly be formulated as}
\begin{equation}\label{eq:compress_rate2}
\rev{\varphi_{QABBA}  = \frac{8N + (B_{\texttt{len}} + B_{\texttt{inc}}) \cdot k +  p\cdot B_T + 4 B_s}{B_T \cdot n }.}
\end{equation}

\rev{If we use $p=1, B_T=32, B_{\texttt{len}}=8, B_{\texttt{inc}}=12$, and $B_s=32$ for \eqref{eq:compress_rate2}, then we have $\varphi_{QABBA}  = \frac{ 8N + 20 \cdot k + 160}{32 \cdot n}$.}

\rev{The byte-symbol term $8N$ can be further generalized when the symbolic sequence is encoded by an optional lossless second-stage codec. Let $\Gamma$ be an invertible encoder applied to the symbol string $A=[a_1,\ldots,a_N]\in\mathcal{L}^N$, and let $B_{\Gamma}(A)$ denote the number of bits required by the encoded sequence, including any codec metadata needed for exact decoding. Since $\Gamma$ is lossless, decoding recovers the same symbolic sequence $A$ and therefore does not change the QABBA reconstruction error. Under this optional post-processing, the QABBA storage ratio becomes}
\begin{equation}\label{eq:compress_rate_gamma}
\rev{\varphi_{QABBA,\Gamma}  = \frac{B_{\Gamma}(A) + (B_{\texttt{len}} + B_{\texttt{inc}}) \cdot k +  p\cdot B_T + 4 B_s}{B_T \cdot n }.}
\end{equation}
\rev{The original byte-symbol model in \eqref{eq:compress_rate2} is recovered by setting $B_{\Gamma}(A)=8N$. Practical choices for $\Gamma$ include fixed-width alphabet-aware packing with $\lceil\log_2 |\mathcal{L}|\rceil$ bits per symbol, Huffman coding, and dictionary-based codecs such as LZW. This second-stage codec is not part of the lossy approximation itself; it is an orthogonal storage layer applied after compression, digitization, and quantization.}

\rev{As explained in \eqref{eq:compress_rate1}--\eqref{eq:compress_rate_gamma}, the storage ratio of the ABBA representation depends on $N, k$, $n$, and the coded length of the symbolic sequence, which are manageable by tuning the hyperparameters $\texttt{tol}$, $k$, and $\alpha$ to obtain an acceptable approximation error. The storage efficiency of QABBA over ABBA originates from using lower bit-widths to decrease $B_{\texttt{len}}$ and $B_{\texttt{inc}}$, thereby reducing the contribution of the symbolic centers. The required storage for symbolic approximation resulting from QABBA and ABBA depends heavily on the size of symbolic centers $C$ and the length of the symbolic representation, i.e., $N$.}

\section{Empirical results}\label{sec:exp}

\rev{In this section, we compare the performance of non-quantized ABBA methods, i.e., ABBA and fABBA, quantized ABBA, and SAX-based algorithms in terms of storage and approximation errors. We conduct extensive experiments on synthetic time series as well as well-established real-world datasets, including those from the UCR Archive \cite{UCRArchive2018} and the UEA Archive \cite{UEAArchive2018}.} To ensure a fair comparison, we selected competing algorithms with publicly available software implementations.\footnote{ABBA and fABBA are accessible at \url{https://github.com/nla-group/ABBA} and \url{https://github.com/nla-group/fABBA}, respectively.} \rev{The integer arithmetic inside QABBA is emulated using Python built-in integer types. Moreover, our software is integrated into the established fABBA library} (\url{https://github.com/nla-group/fABBA}), and the experimental code is openly available at:
\begin{center}
    \url{https://github.com/inEXASCALE/qabba}.
\end{center}

In the following experiments, we compare QABBA with ABBA, fABBA, SAX, and $1d$-SAX algorithms.  Additionally, we test two variants of QABBA{---}QABBA (VQ) and QABBA (GA){---}which use \texttt{k-means++} \cite{1283494} and greedy aggregation \cite{fABBA2022}, respectively, in the digitization stage. All experiments here except Section~\ref{sec:llm} are simulated in {64-bit} Python on a Dell PowerEdge R740 Server with Intel Xeon Silver @ 4114 2.2GHz and 1.5 TB RAM.

\subsection{Approximation error from quantization}

To evaluate QABBA, it is critical to understand how the number of bits used affects the performance of symbolic approximation. In this experiment, we evaluate the performance of QABBA in terms of the number of bits used to store the lengths and increments. The experimental data are 100 synthetic time series of length 5,000 generated using a standard normal Gaussian distribution. For QABBA, we quantize either \texttt{len} or \texttt{inc} to a lower bit-width integer format while fixing the other to float32 format. The approximation error is evaluated in terms of the MSE (\eqref{eq:mse}) distance between the original time series and the reconstructed time series.

\begin{figure}[ht!]
\centering
\subfloat[\texttt{len}]{\includegraphics[width=0.322\textwidth]{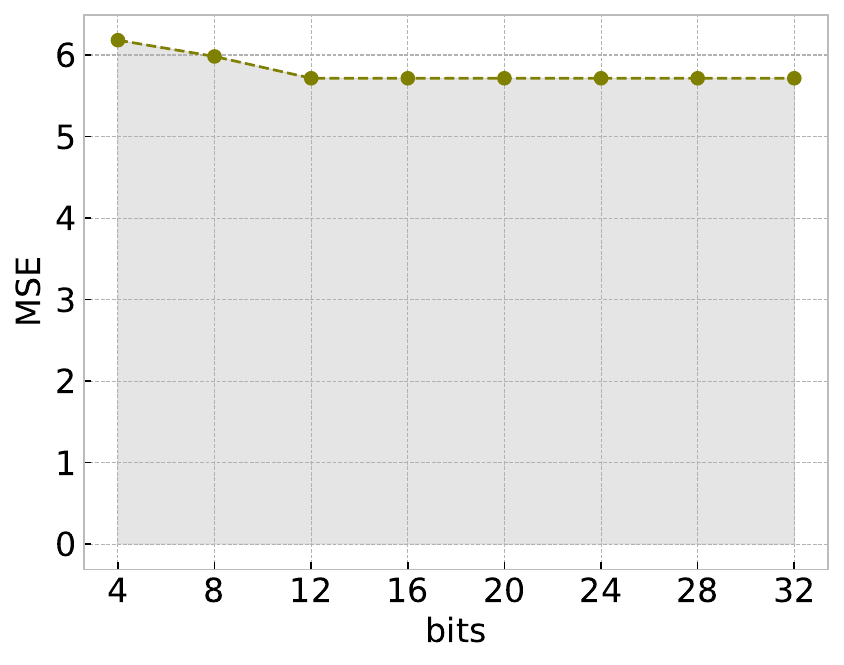}}
\subfloat[\texttt{inc}]{\includegraphics[width=0.345\textwidth]{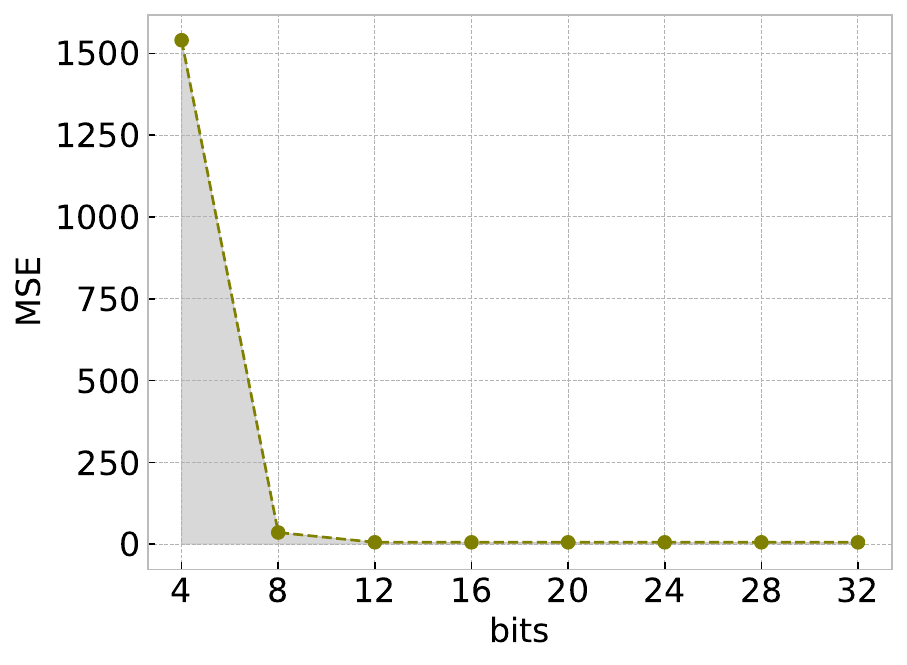}}
\caption{Reconstruction error arising from quantization.}
\label{fig:qerr}
\end{figure}

The results are depicted in \figurename~\ref{fig:qerr}. We can see that the error arising from quantizing the increment, compared to the \texttt{len}, increases dramatically as the bit-width decreases. In other words, the impact of quantizing the \texttt{len} values is much less than that of quantizing the \texttt{inc} values. The safety margin of the quantization bits for \texttt{len} quantization in this dataset is in $[8, 32)$ whereas for \texttt{inc} quantization it is $[12, 32)$ (for there to be an advantage to quantization, the bit-width should be lower than 32).
This provides guidance for choosing the bit-width for the quantizations in practice. All experiments in the following sections will simulate QABBA with a quantization of 8 bits for the \texttt{len} values of the symbolic centers and 12 bits for the \texttt{inc} values of the symbolic centers.

\subsection{LLM with QABBA for time series regression}\label{sec:llm}

\begin{table}[htp!]
\centering
\caption{Hyperparameters of the time series regression task. Quant. is the LoRA quantization process. Inhib. is the inhibition threshold in QLoRA. MAE is mean absolute error. }
\label{TableHyper}\setlength{\extrarowheight}{1pt}
\scriptsize\setlength\tabcolsep{1.8pt}
\begin{tabular}{ccccc|cccccc}
\hline
\multicolumn{5}{c|}{QLoRA Hyperparameters} & \multicolumn{6}{c}{QABBA Hyperparameters}\\ \hline
\begin{tabular}[c]{@{}c@{}}quant\\ 4-bit\end{tabular} & \begin{tabular}[c]{@{}c@{}}low\\ rank\end{tabular} & $\alpha$ & inhi. & dropout & \texttt{inc}    & \texttt{len}           & init & \texttt{tol}   & $\alpha$ & \texttt{scl}\\ \hline
True                                                  & 16   & 16    & 0.3   & 0.05    & 16, 32 & 32, 16, 12, 8 & agg  & 0.05 & 0.05 & 3
\\ \hline
\end{tabular}
\vspace{7pt}
\begin{tabular}{cccccccc}
\multicolumn{8}{c}{LLM Hyperparameters} \\[1mm] \hline
\begin{tabular}[c]{@{}c@{}}batch\\ size\end{tabular} & \begin{tabular}[c]{@{}c@{}}learning\\ rate\end{tabular} & epochs & \begin{tabular}[c]{@{}c@{}}max\\ length\end{tabular} & optim & \begin{tabular}[c]{@{}c@{}}weight\\ decay\end{tabular}    & \begin{tabular}[c]{@{}c@{}}warmup\\ steps\end{tabular}           & \begin{tabular}[c]{@{}c@{}}loss\\ function\end{tabular} \\ \hline
4  & 2e-4   & 20    & 4,096   & adamw\_8bit    & 5e-5  & 5  & MAE
\\ \hline
\end{tabular}
\end{table}

\begin{table}[htp!]
\centering
\caption{RMSE performance of time series regression tasks on QABBA (Part 1). SOTA results are from \cite{tan2021time}. Best results are in \textbf{bold}.}
\label{tab:sota_part1}
\scriptsize
\setlength{\tabcolsep}{3pt} 
\begin{tabular}{lcc *{8}{S[table-format=3.2]}}
\toprule
\multirow{2}{*}{LLM} & \multicolumn{2}{c}{Quantize} & \multicolumn{8}{c}{Monash Regression Data} \\
\cmidrule(lr){2-3} \cmidrule(lr){4-11}
& \texttt{len} & \texttt{inc} & {App.} & {Hou.1} & {Hou.2} & {Ben.} & {Bei.1} & {Bei.2} & {Liv.} & {Flo.1} \\
\midrule
Mistral-7B & 32 & 32 & 2.34 & 228.83 & 24.51 & 4.03 & 65.24 & 53.49 & 20.88 & 0.36 \\
Mistral-7B & 16 & 16 & 2.34 & 228.63 & \textbf{24.51} & 4.03 & \textbf{65.24} & 53.49 & \textbf{20.85} & 0.36 \\
Mistral-7B & 12 & 16 & 2.34 & 228.78 & 24.56 & 4.03 & 65.25 & \textbf{53.49} & 20.88 & 0.36 \\
Mistral-7B & 8  & 16 & 2.43 & 228.83 & 24.54 & 4.03 & 65.25 & 53.50 & 20.94 & 0.37 \\
\midrule
SOTA       & -- & -- & \textbf{2.29} & \textbf{132.80} & 32.61 & \textbf{0.64} & 93.14 & 59.50 & 28.80 & \textbf{0.00} \\
\bottomrule
\end{tabular}
\end{table}

\begin{table}[htp!]
\centering
\caption{RMSE performance of time series regression tasks on QABBA (Part 2). SOTA results are from \cite{tan2021time}. Best results are in \textbf{bold}.}
\label{tab:sota_part2}
\scriptsize
\setlength{\tabcolsep}{3pt} 
\begin{tabular}{lcc *{8}{S[table-format=2.2]}}
\toprule
\multirow{2}{*}{LLM} & \multicolumn{2}{c}{Quantize} & \multicolumn{8}{c}{Monash Regression Data} \\
\cmidrule(lr){2-3} \cmidrule(lr){4-11}
& \texttt{len} & \texttt{inc} & {Flo.2} & {Flo.3} & {Aus.} & {PPG.} & {IEE.} & {Ne.H} & {Ne.T} & {Cov.} \\
\midrule
Mistral-7B & 32 & 32 & 0.39 & 0.39 & 5.89 & 12.58 & 23.03 & 0.10 & 0.10 & 0.03 \\
Mistral-7B & 16 & 16 & 0.39 & 0.37 & \textbf{5.89} & 12.58 & \textbf{23.02} & \textbf{0.10} & 0.10 & 0.03 \\
Mistral-7B & 12 & 16 & 0.39 & 0.39 & 5.90 & 12.61 & 23.05 & 0.11 & \textbf{0.10} & 0.03 \\
Mistral-7B & 8  & 16 & 0.40 & 0.41 & 5.92 & 12.62 & 23.08 & 0.12 & 0.11 & \textbf{0.03} \\
\midrule
SOTA       & -- & -- & \textbf{0.01} & \textbf{0.00} & 8.12 & \textbf{9.92} & 23.90 & 0.14 & 0.14 & 0.04 \\
\bottomrule
\end{tabular}
\end{table}

\rev{In this section, we show that QABBA can expose symbolic temporal patterns to LLMs for time-series regression. For the regression task, we evaluate one pretrained LLM using the Time Series Extrinsic Regression (TSER) benchmark archive \cite{tan2021time}, which contains 19 time-series datasets from five distinct areas: health monitoring, energy monitoring, environmental monitoring, sentiment analysis, and forecasting\footnote{Monash regression data is available at \url{http://tseregression.org/}.}.} Details of the hyperparameter settings of our implementation can be found in \tablename~\ref{TableHyper}. \rev{We simulated QABBA under various bit-width settings for $\len$ and $\inc$ and applied a commonly used LLM, Mistral-7B~\cite{jiang2023mistral}, to time-series regression problems. In the first step, each numerical time series is transformed into a symbolic series, while the corresponding label remains numerical. By replacing the last layer with a regression layer, we fine-tune Mistral-7B on the training data using QLoRA \cite{dettmers2024qlora} with inhibition \cite{kang2024ina}.} All the experiments are simulated in PyTorch \cite{paszke2019pytorch} and conducted on a single NVIDIA A100 40GB GPU.

\rev{The empirical results are shown in \tablename~\ref{tab:sota_part1} and \tablename~\ref{tab:sota_part2}; QABBA outperforms the machine-learning state-of-the-art (SOTA) results in 9 out of 16 use cases. Specifically, when using a 16-bit width for both $\texttt{len}$ and $\texttt{inc}$, QABBA shows competitive performance for time-series regression while using less memory. We also observe that lower bit widths for $\texttt{len}$ and $\texttt{inc}$ do not substantially degrade regression performance. QABBA provides chains of patterns to LLMs by compressing and digitizing time series into symbols, which changes the embedding space through adaptation-based fine-tuning methods. The benefits of QABBA with LLMs include (1) avoiding the need for LLMs to learn time-series embeddings from scratch and (2) using only compression and decompression without training extra embeddings \cite{jin2023time}.}

\subsection{Performance profiles in UCR Archive}\label{sec:ucr}
\rev{The concept of \emph{performance profiles}~\cite{Dolan2002Benchmarking} has been introduced as an efficient benchmarking tool for evaluating a set of optimization algorithms over a large set of test problems. Performance profiles reduce the influence of a small number of atypical problems and are less sensitive to rank-based comparisons of algorithms.}

The performance profile for an algorithm is represented by the cumulative distribution function (CDF) for a performance measure. Let $\mathcal{V}$ be a set of problems and let $\mathcal S$ be a set of algorithms. The quantity $s_{i,j} \ge 0$ denotes a performance measurement of algorithm $i \in \mathcal S$ run on problem $j \in \mathcal{V}$, assuming that a smaller value of $s_{i,j}$ is better. For any problem $j \in \mathcal{V}$, let $\widehat{s}_j = \min\{s_{i,j}:i \in \mathcal S\}$ be the best performance of an algorithm on problem $j$. The performance ratio $r_{i,j}$ is given by
\begin{equation}
	r_{i,j} = \frac{s_{i,j}}{\widehat{s}_{j}},
\end{equation}
\rev{If algorithm $i$ fails on problem~$j$, $r_{i,j}$ is conventionally set to $+\infty$.}
We define
\begin{equation}
	\xi(r_{i,j}, \theta) = \left\{
	\begin{array}{rcl}
		1 & & \text{if} \hspace{5pt} {r_{i,j} < \theta}\\
		0 & & \text{otherwise}\\
	\end{array} \right.
\end{equation}
where $\theta \ge 1$. The \emph{performance profile of solver~$i$} is
\begin{equation}
	 \rho_{i}(\theta) = \frac{\sum_{j \in \mathcal{V}}\xi(r_{i,j}, \theta)}{|\mathcal{V}|},\quad  \theta \ge 1.
\end{equation}

From a statistics perspective, $\rho_{i}(\theta)$ can be regarded as the empirical probability that the performance ratio $r_{i,j}$ for algorithm~$i \in \mathcal S$ and each problem $j \in \mathcal{V}$ is within a factor $\theta$ of the best possible ratio. The graph of the CDF $\rho_{i}${---}the performance profile{---}reveals several features in a visually convenient way, e.g., a line which more closely approaches the upper left corner indicates a better algorithm. In particular, algorithms~$i$ with fast-increasing performance profiles are preferable if the problem set $\mathcal{V}$ is suitably large and representative of problems likely to occur in applications.

The setting of the first experiment follows \cite{EG19b}. In this simulation, we evaluate SAX, 1d-SAX, ABBA, fABBA, and QABBA (VQ \& GA) across all datasets (a total of 201,161 time series) in the UCR Archive. The experimental design follows \cite{fABBA2022}; we first run fABBA with a given digitization rate $\alpha$, and then take the number of distinct symbols of the representation as $k$ and feed the same $\alpha$ and $k$ values to other symbolic representation methods for a fair comparison. We evaluate Euclidean distance, DTW distance, and their differenced counterparts to benchmark performance on the UCR Archive data in terms of two different settings of digitization rate. Recall that QABBA variants use 8 bits for the lengths and 12 bits for the increments.

\begin{figure}[ht!]
\centering
\subfloat[Euclidean distance.]{\includegraphics[width=0.42\textwidth]{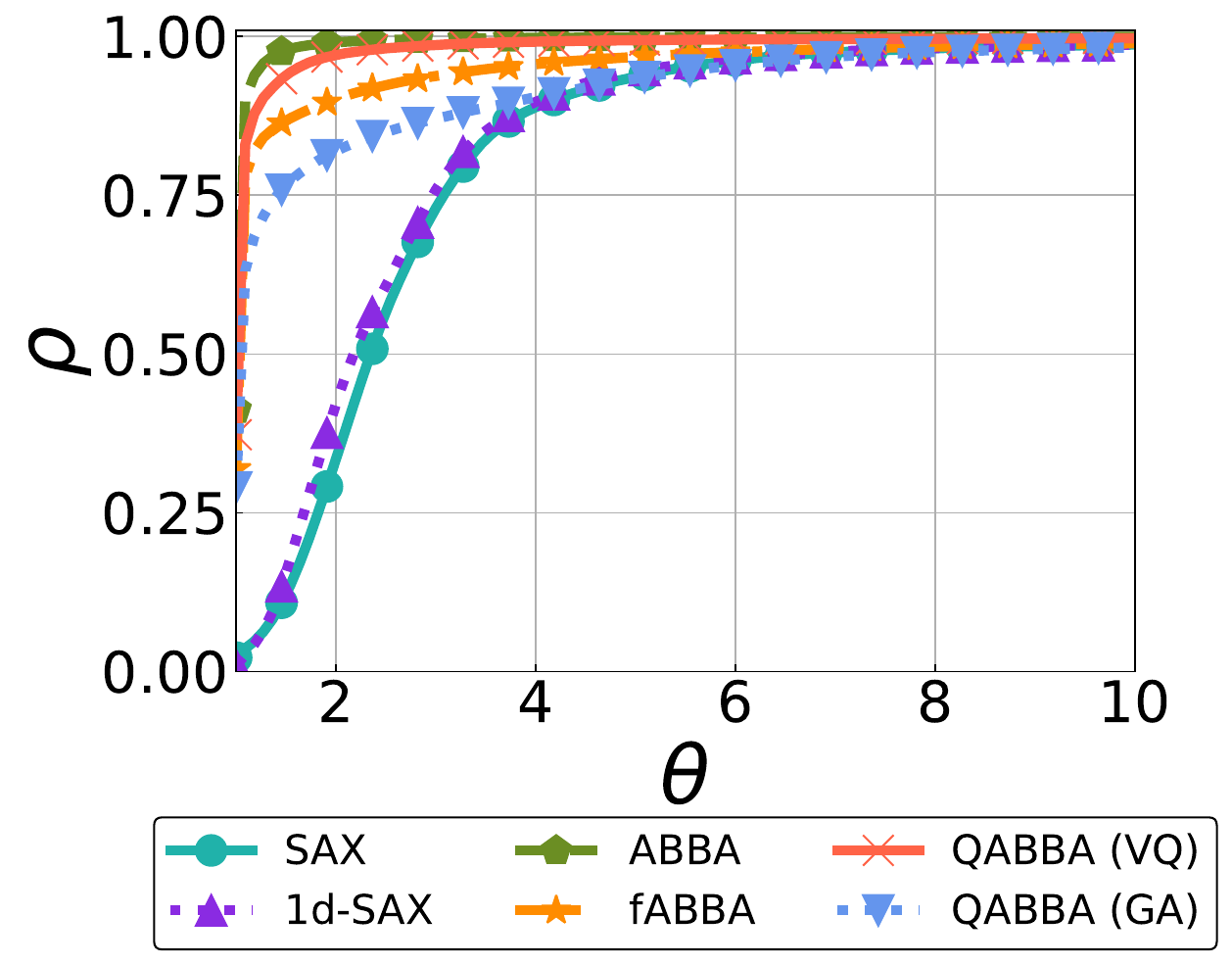}}
\subfloat[DTW distance.]{\includegraphics[width=0.42\textwidth]{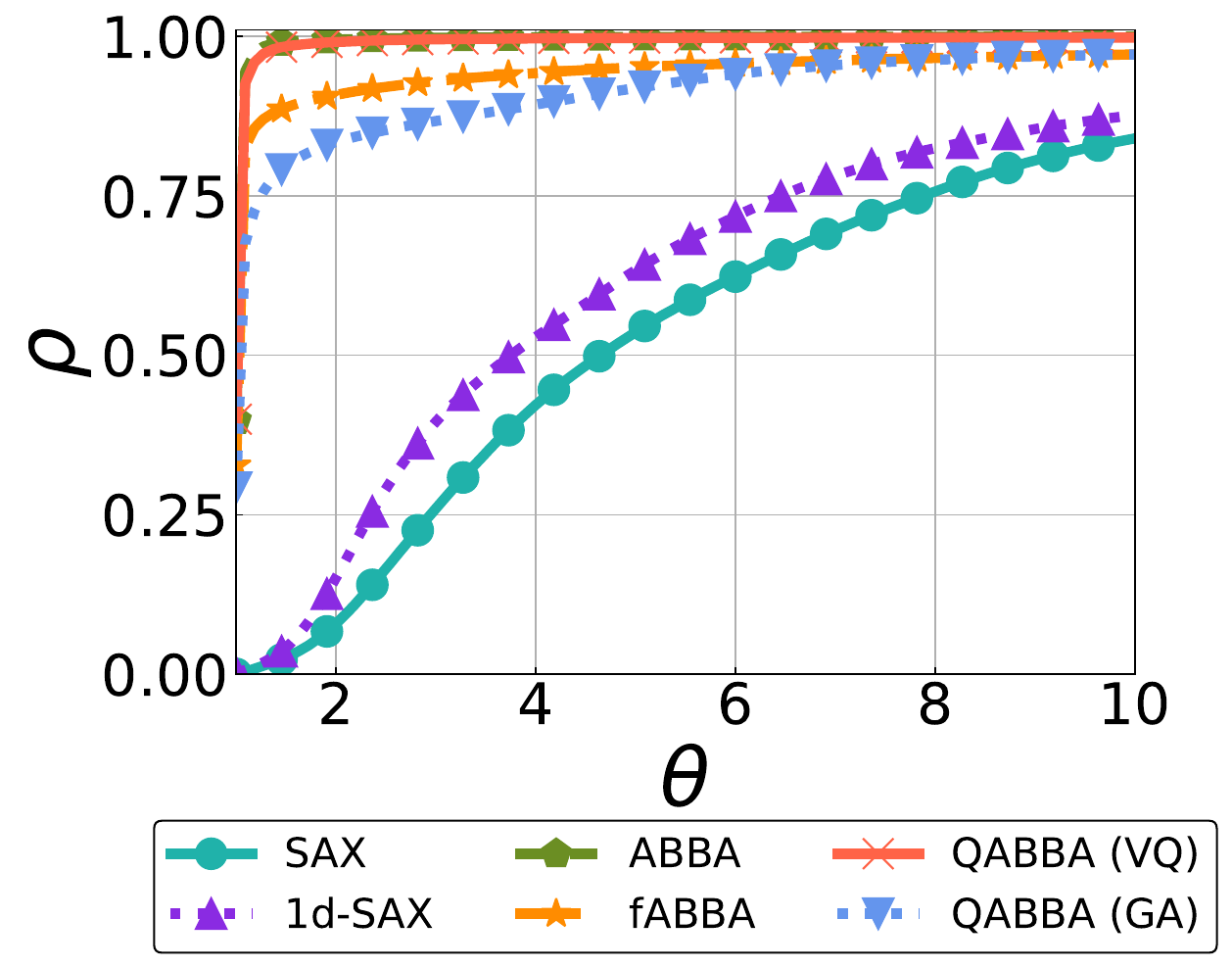}}\\

\subfloat[Euclidean distance (differenced).]{\includegraphics[width=0.42\textwidth]{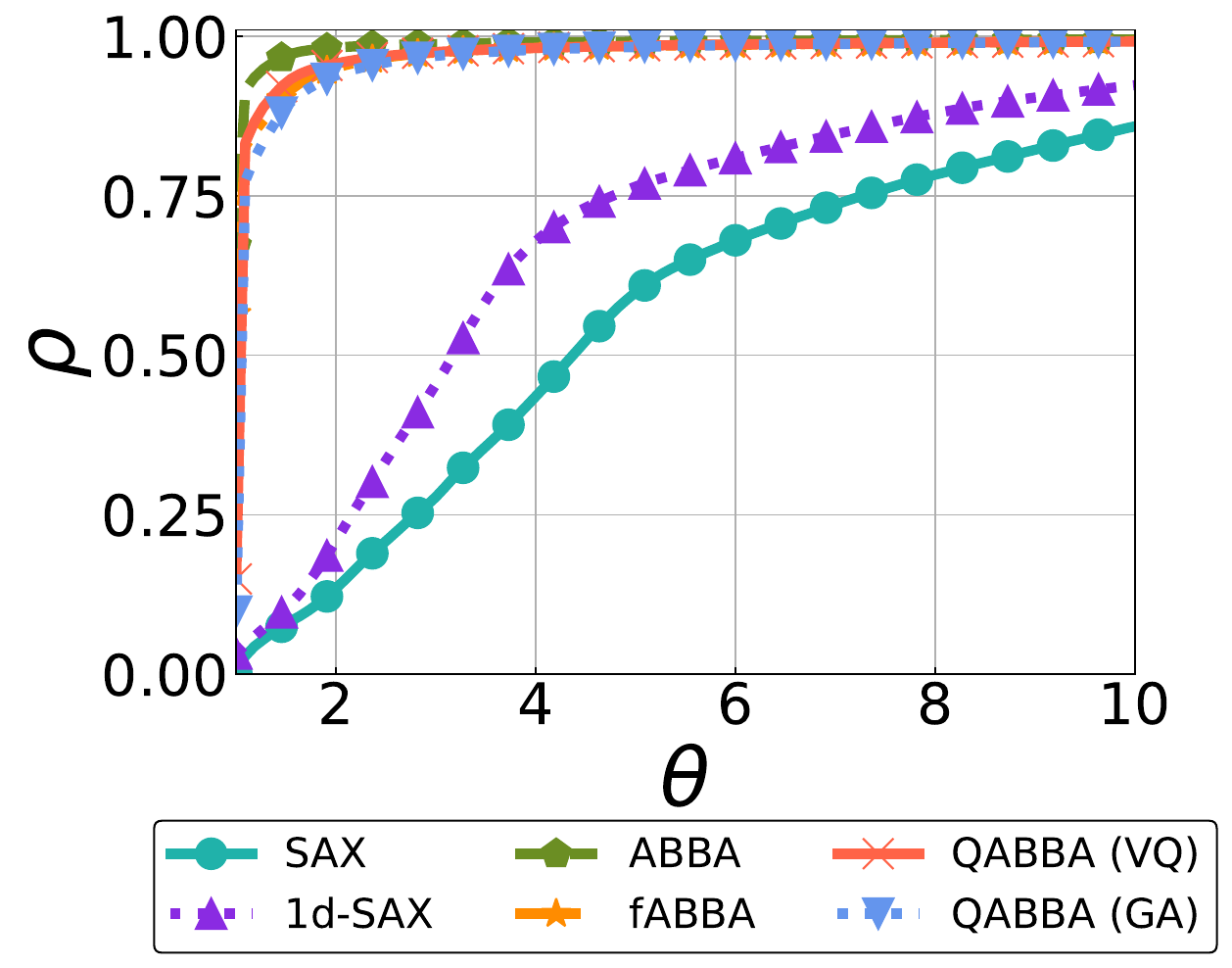}}
\subfloat[DTW distance (differenced).]{\includegraphics[width=0.42\textwidth]{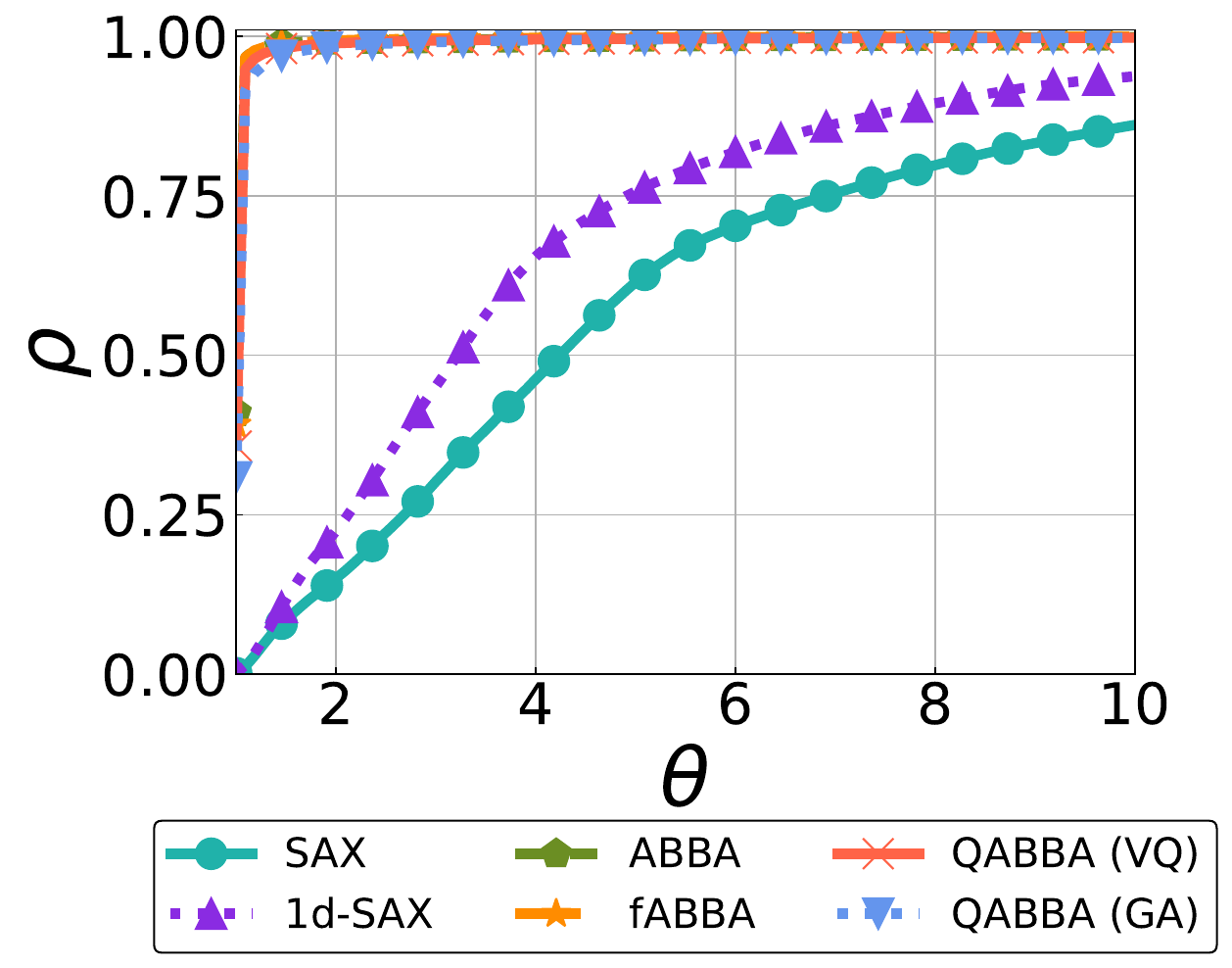}}
\caption{Performance profiles ($\alpha = 0.1$).}
\label{fig:PP0.1}
\end{figure}

\begin{figure}[ht]
\centering
\subfloat[Euclidean distance.]{\includegraphics[width=0.42\textwidth]{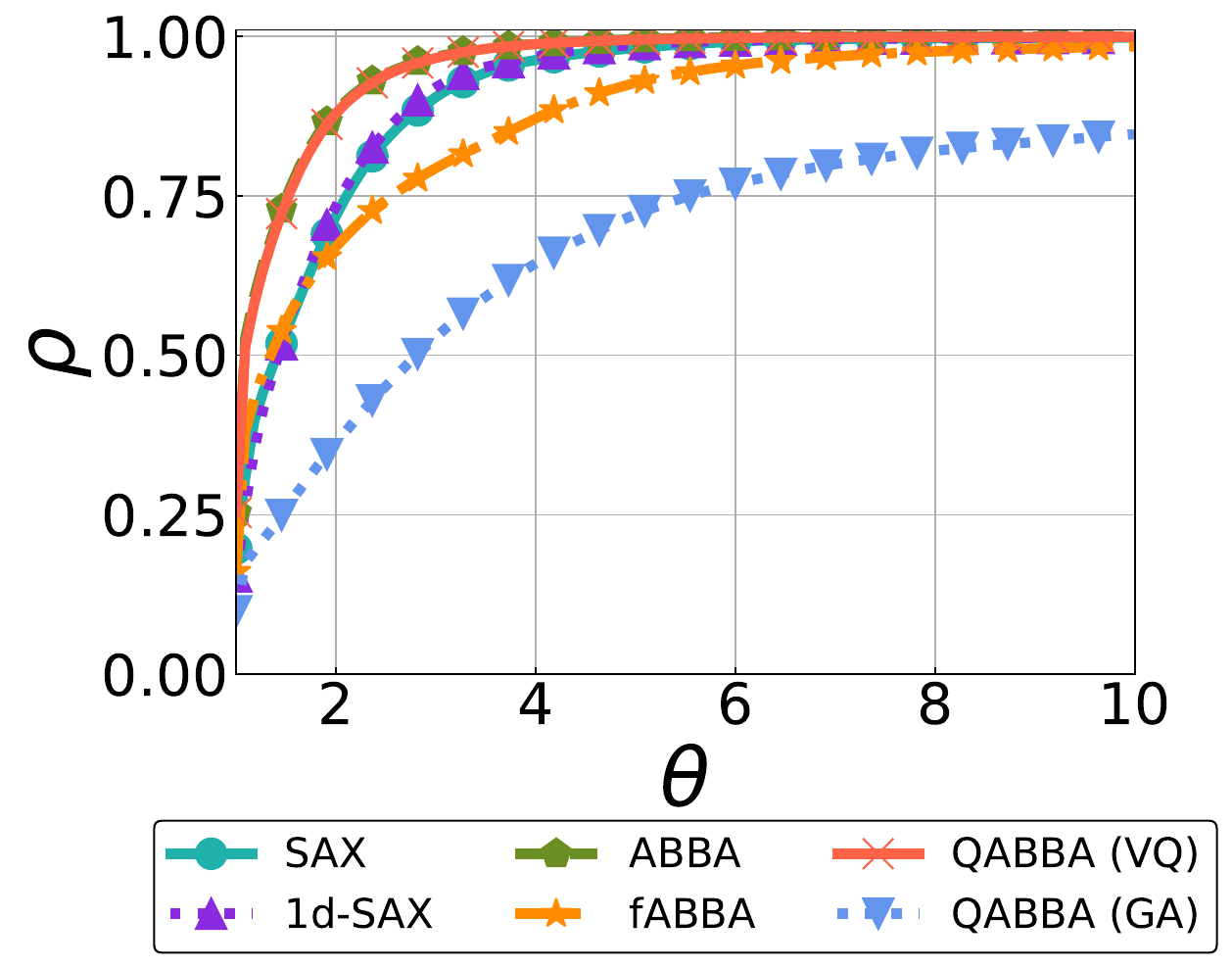}}
\subfloat[DTW distance.]{\includegraphics[width=0.42\textwidth]{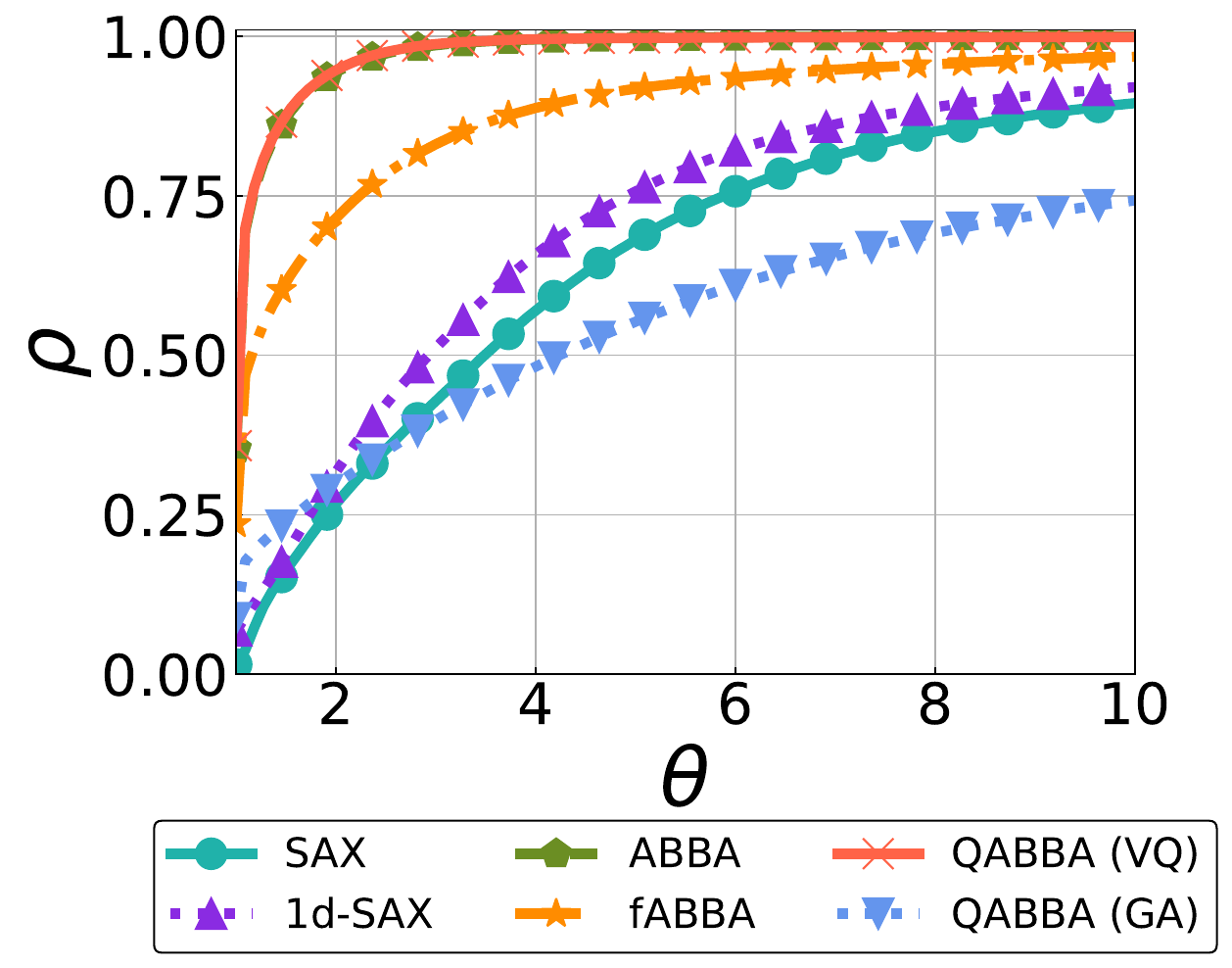}}\\

\subfloat[Euclidean distance (differenced).]{\includegraphics[width=0.42\textwidth]{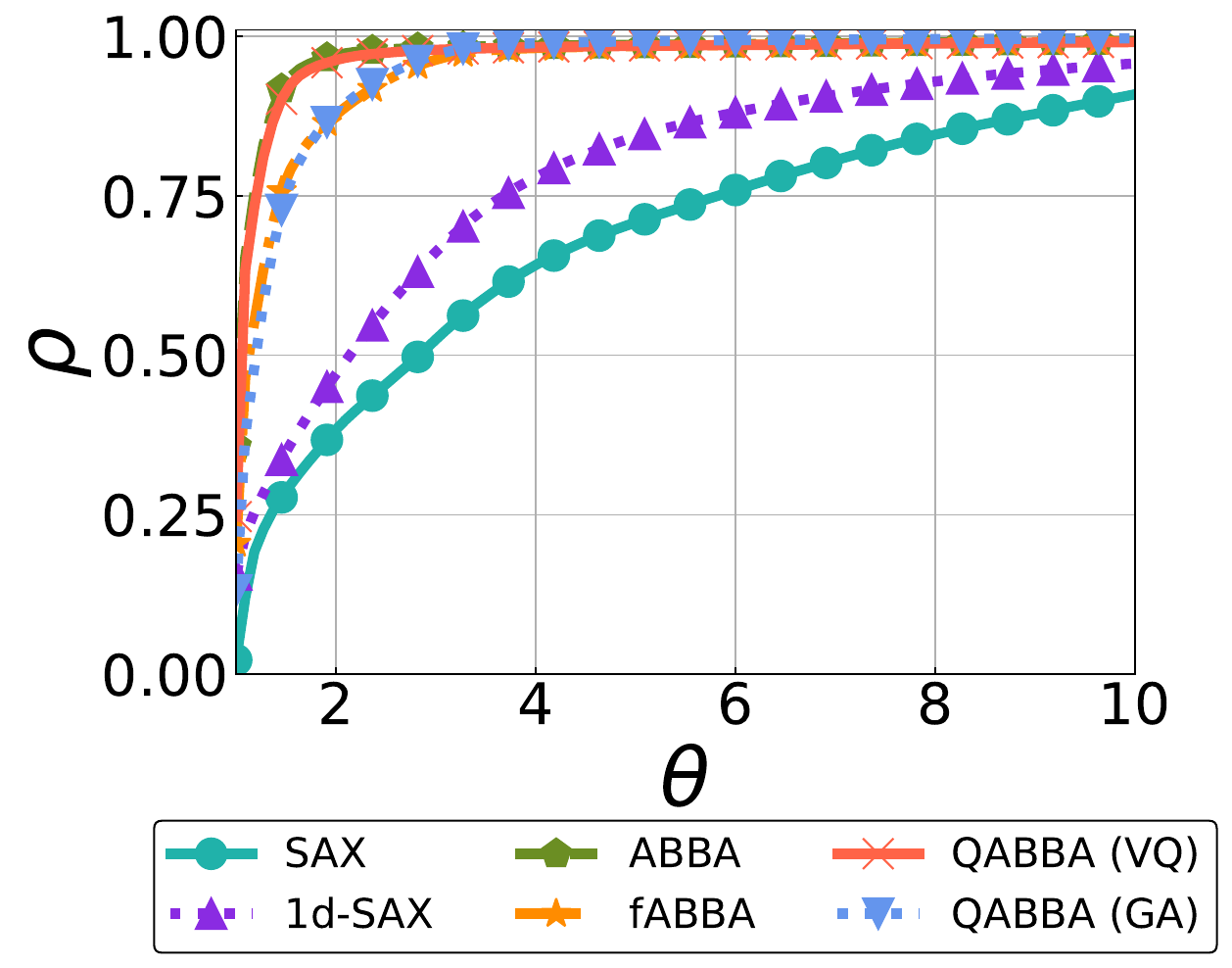}}
\subfloat[DTW distance (differenced).]{\includegraphics[width=0.42\textwidth]{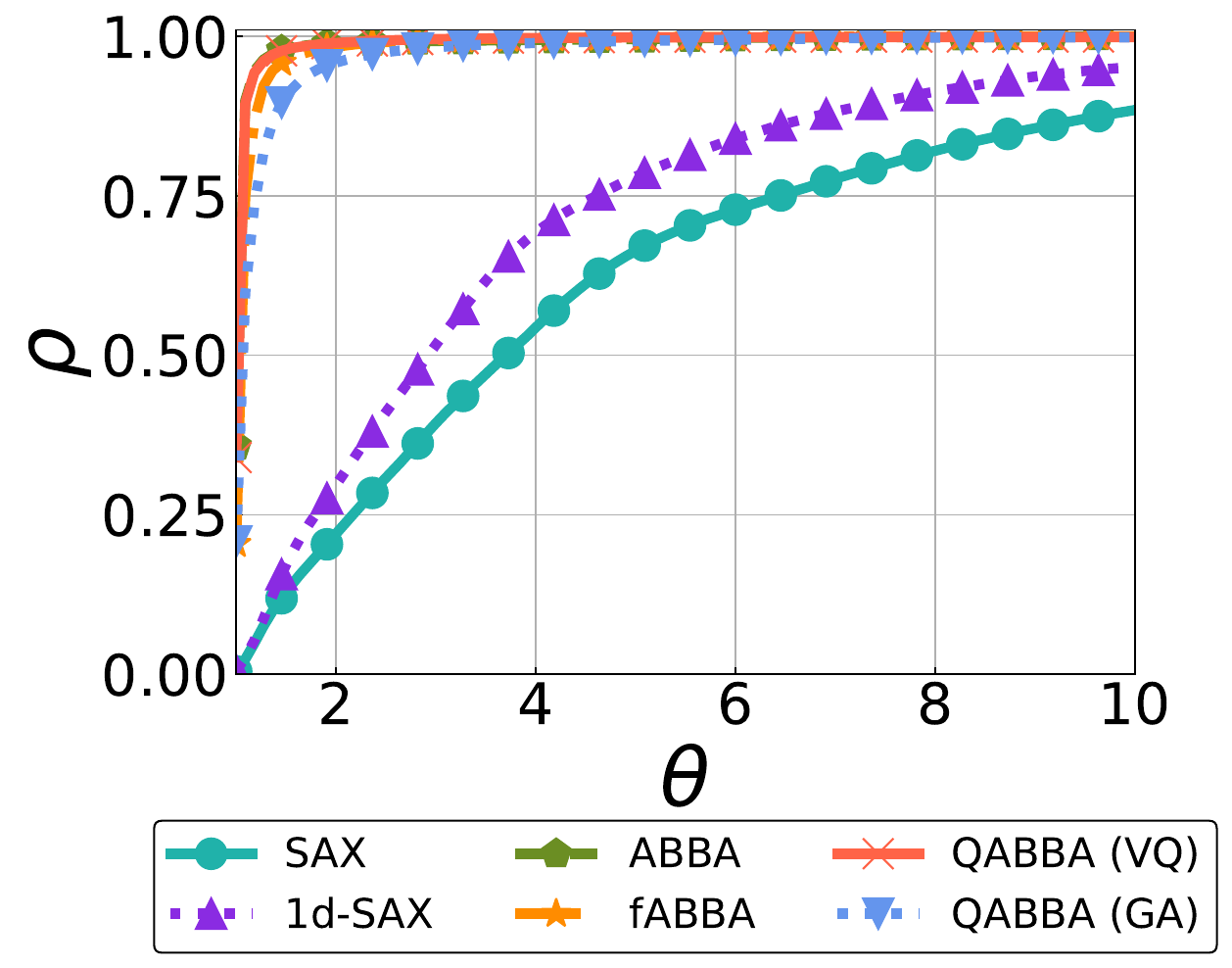}}
\caption{Performance profiles ($\alpha = 0.5$).}
\label{fig:PP0.5}
\end{figure}

\begin{figure}[ht]
	\centering
	\subfloat[$\alpha = 0.1$]{\includegraphics[width=0.42\textwidth]{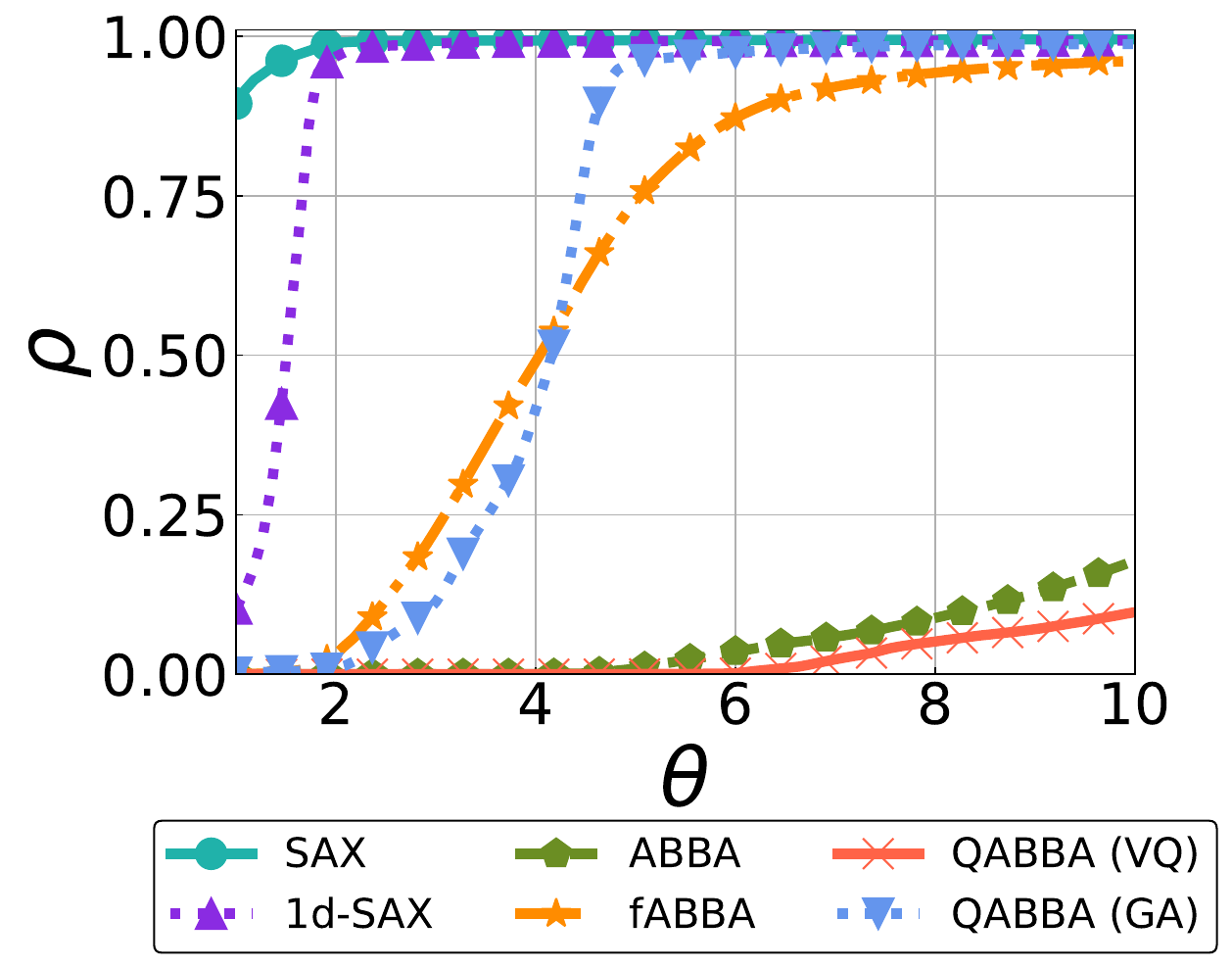}}\subfloat[$\alpha = 0.5$]{\includegraphics[width=0.42\textwidth]{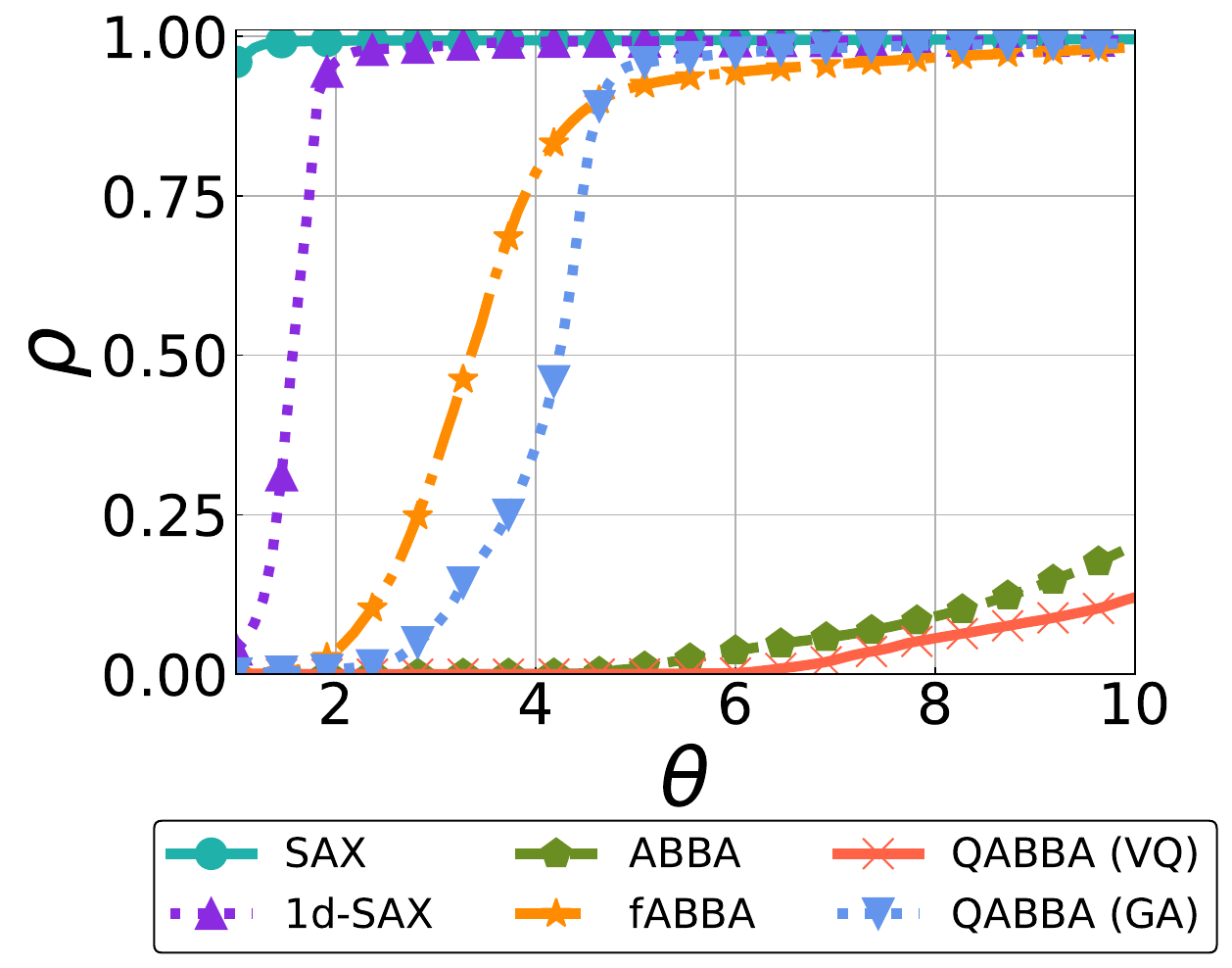}}
	\caption{Performance profiles in Runtime.}
	\label{fig:runtime}
\end{figure}

\rev{The resulting performance profiles are depicted in \figurename~\ref{fig:PP0.1} and \figurename~\ref{fig:PP0.5}. First, we observe that QABBA results in a slight reconstruction loss, and the loss increases with $\alpha$ (i.e., the fewer symbolic centers used, the greater the loss for QABBA reconstruction). Second, QABBA (VQ) closely follows ABBA in both \figurename~\ref{fig:PP0.1} and \figurename~\ref{fig:PP0.5}, especially in \figurename~\ref{fig:PP0.5} with $\alpha=0.5$. Additionally, the runtime comparison between QABBA and ABBA is shown in \figurename~\ref{fig:runtime}, which indicates that quantization introduces almost no speed loss within the ABBA framework. Overall, QABBA exhibits reconstruction performance comparable to ABBA.}

\subsection{Storage Reduction and Recovery Quality of Time Series}\label{sec:uea}

{
\rev{In this section, we discuss how much storage reduction can be achieved in practice. We evaluate QABBA and its predecessors ABBA and fABBA in terms of approximation error and storage ratio for selected datasets from the Monash regression data (see \tablename~\ref{TableMemory}) and UEA Archive (see \tablename~\ref{tab:UEAinfo}).}

\rev{Compared with uploading raw data, ABBA-family methods can reduce the amount of uploaded data by orders of magnitude, which can substantially improve communication efficiency in bandwidth-limited settings. Moreover, QABBA reduces memory storage by quantizing the symbolic centers; additional reductions are possible when the strings are compressed by the LZW algorithm. We do not report computing-speed measurements in this experiment because the Python implementation emulates integer arithmetic and does not expose the low-level hardware behavior needed for a fair speed comparison.}

\begin{table}[htp!]
\centering
\caption{{The number of bits when uploading QABBA parameters from the IoT devices to the LLM environment.} }
\label{TableMemory}\setlength{\extrarowheight}{1pt}
\scriptsize\setlength\tabcolsep{6.3pt}
\begin{tabular}{cccccccc}
\hline
\multirow{2}{*}{Data}        & \multicolumn{2}{c}{Quantize} & \multicolumn{4}{c}{Uploaded Items}                                                                                                & \multirow{2}{*}{Total} \\ \cmidrule(lr){2-3} \cmidrule(lr){4-7}
                                      &  \texttt{len}     &  \texttt{inc}     & \texttt{alphabet set} & \texttt{centers} & \texttt{start set} & \texttt{strings} &                                 \\ \hline
\multirow{4}{*}{\rotatebox{90}{Cov.}}
& 32           & 32           & 240$\times$8               & 240$\times$(32+32) & 140$\times$32           & 1,009$\times$8             & $29,832\thickapprox2^{15}$    \\
& 16           & 16           & 240$\times$8               & 240$\times$(16+16) & 140$\times$16           & 1,009$\times$8             & $19,912\thickapprox2^{14}$    \\
& 12           & 16           & 240$\times$8               & 240$\times$(12+16) & 140$\times$12           & 1,009$\times$8             & $18,392\thickapprox2^{14}$    \\
& 8           & 16           & 240$\times$8               & 240$\times$(8+16) & 140$\times$8           & 1,009$\times$8             & $16,872\thickapprox2^{14}$    \\
 \hline
\multirow{4}{*}{\rotatebox{90}{IEE.}}
& 32           & 32           & 5,088$\times$16               & 5,088$\times$(32+32) & 1,768$\times$32           & 4,035,422$\times$16             & $65,030,368\thickapprox2^{26}$    \\
& 16           & 16           & 5,088$\times$16               & 5,088$\times$(16+16) & 1,768$\times$16           & 4,035,422$\times$16             & $64,839,264\thickapprox2^{26}$    \\
& 12           & 16           & 5,088$\times$16               & 5,088$\times$(12+16) & 1,768$\times$12           & 4,035,422$\times$16             & $64,811,840\thickapprox2^{26}$    \\
&8           & 16           & 5,088$\times$16               & 5,088$\times$(8+16) & 1,768$\times$8           & 4,035,422$\times$16             & $64,784,416\thickapprox2^{26}$    \\
 \hline
 \multirow{4}{*}{\rotatebox{90}{PPG.}}
& 32           & 32           & 11,149$\times$16               & 11,149$\times$(32+32) & 43,215$\times$32           & 20,393,699$\times$16             & $328,573,984\thickapprox2^{28}$    \\
& 16           & 16           & 11,149$\times$16               & 11,149$\times$(16+16) & 43,215$\times$16           & 20,393,699$\times$16             & $327,525,776\thickapprox2^{28}$    \\
& 12           & 16           & 11,149$\times$16               & 11,149$\times$(12+16) & 43,215$\times$12           & 20,393,699$\times$16             & $327,308,320\thickapprox2^{28}$    \\
& 8           & 16           & 11,149$\times$16               & 11,149$\times$(8+16) & 43,215$\times$8           & 20,393,699$\times$16             & $327,090,864\thickapprox2^{28}$    \\
 \hline
\end{tabular}
\end{table}

\rev{When uploading the QABBA parameters from IoT devices to the LLM environment, three items are necessary to analyze and recover the time series: the initial value(s), symbolic center(s), and strings. The alphabet set $\mathcal{L}$ is also required so that the receiver can interpret the symbolic vocabulary. For example, the ``Covid3Month'' data contains $140$ samples ($51696$ bytes $=$ $413568\thickapprox2^{19}$ bits), whereas after QABBA processing with $\texttt{len}=8$ and $\texttt{inc}=16$, the uploaded representation uses only $16872\thickapprox2^{14.04}$ bits. In \tablename~\ref{TableMemory}, the number of bits used for the alphabet set $\mathcal{L}$ and the strings is set adaptively according to the cardinality of $\mathcal{L}$. If the cardinality of $\mathcal{L}$ is at most $2^{8}$ (e.g., on the ``Covid3Month'' data), 8 bits are sufficient for each symbol; if the cardinality approaches $2^{16}$, 16 bits are used.}

\begin{figure}[ht!]
\centering
\subfloat[AtrialFibrillation]{\includegraphics[width=0.47\textwidth]{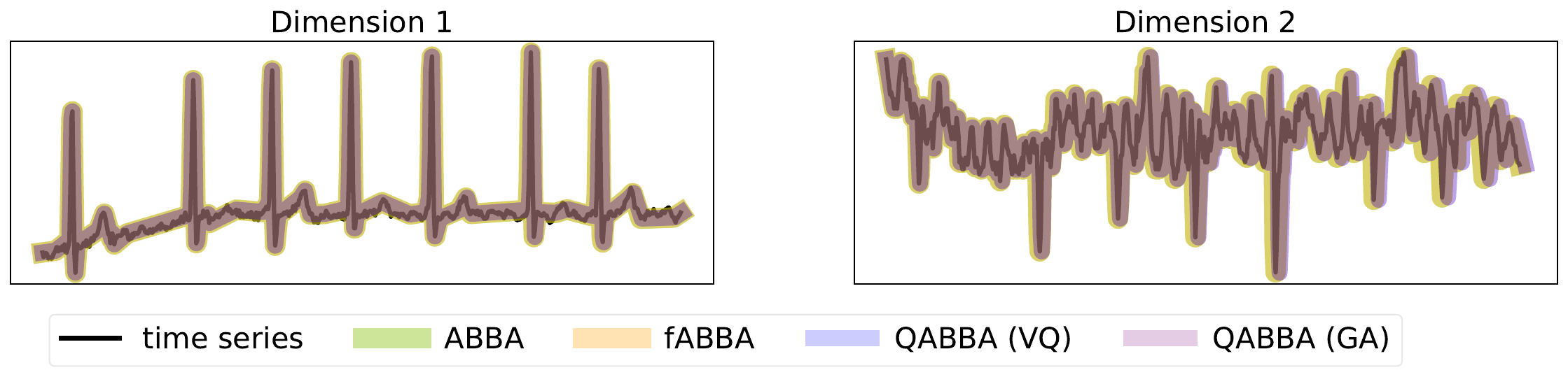}}\hfill
\subfloat[CharacterTrajectories]{\includegraphics[width=0.47\textwidth]{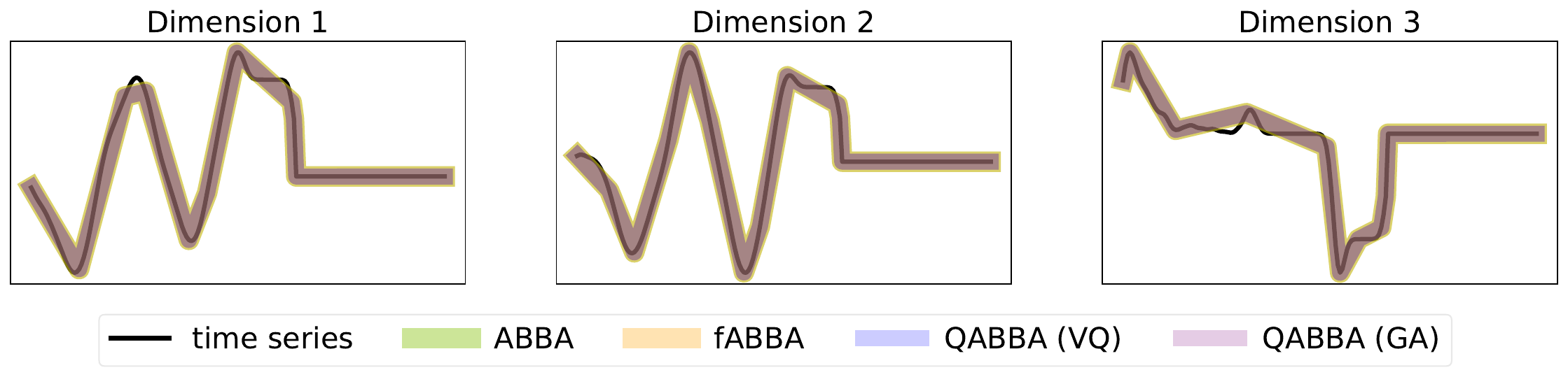}}\\[0.5em]
\subfloat[UWaveGestureLibrary]{\includegraphics[width=0.47\textwidth]{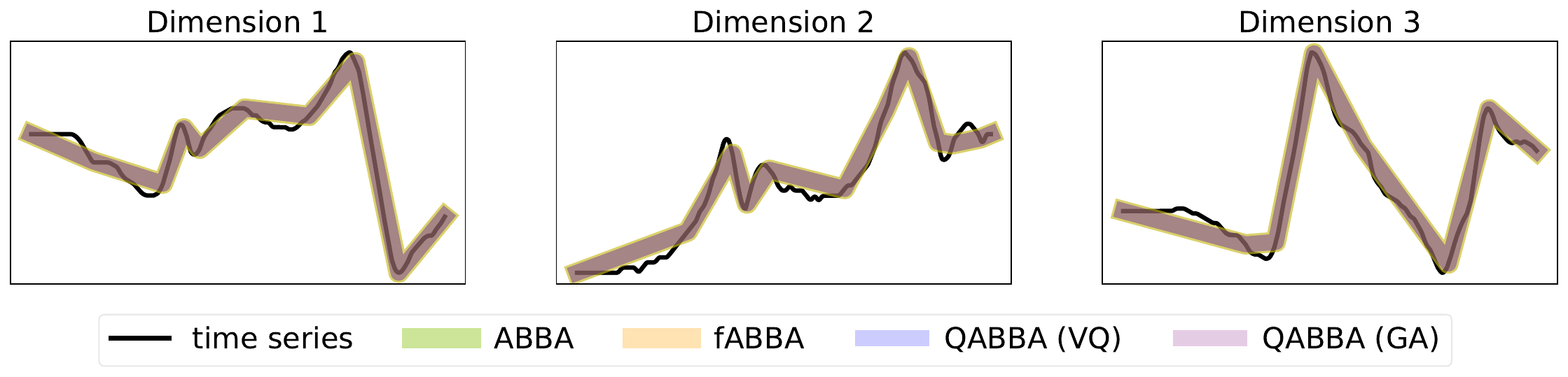}}\hfill
\subfloat[Epilepsy]{\includegraphics[width=0.47\textwidth]{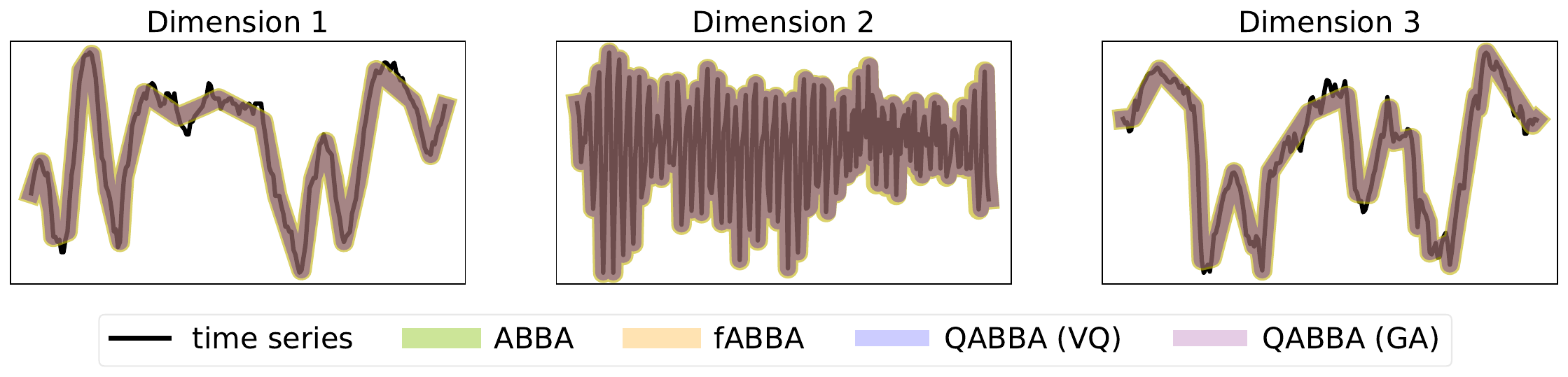}}
\caption{Symbolic reconstruction of UEA Archive (I).}
\label{fig:reconstUEA1}
\end{figure}

\begin{figure}[ht!]
\centering
\subfloat[BasicMotions]{\includegraphics[width=0.47\textwidth]{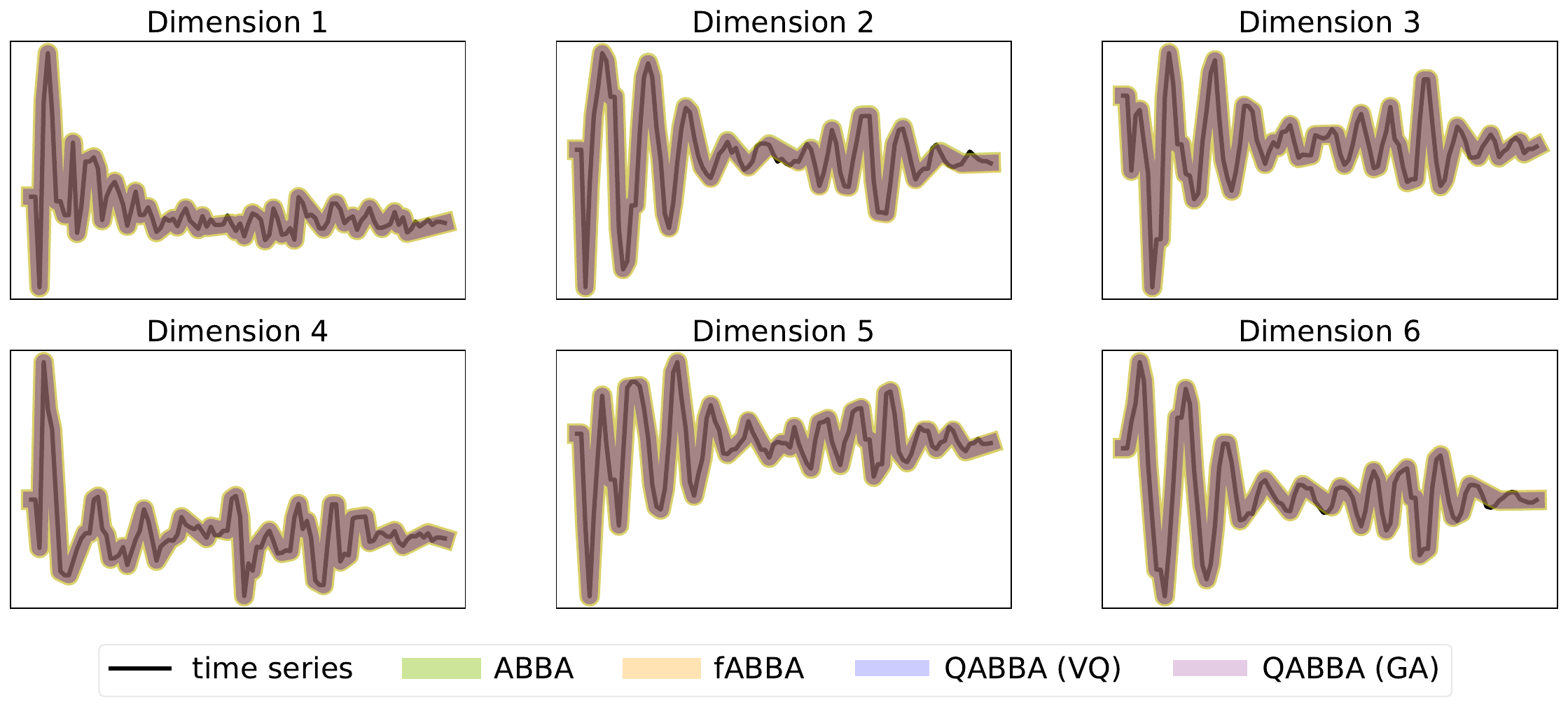}}\hfill
\subfloat[JapaneseVowels]{\includegraphics[width=0.47\textwidth]{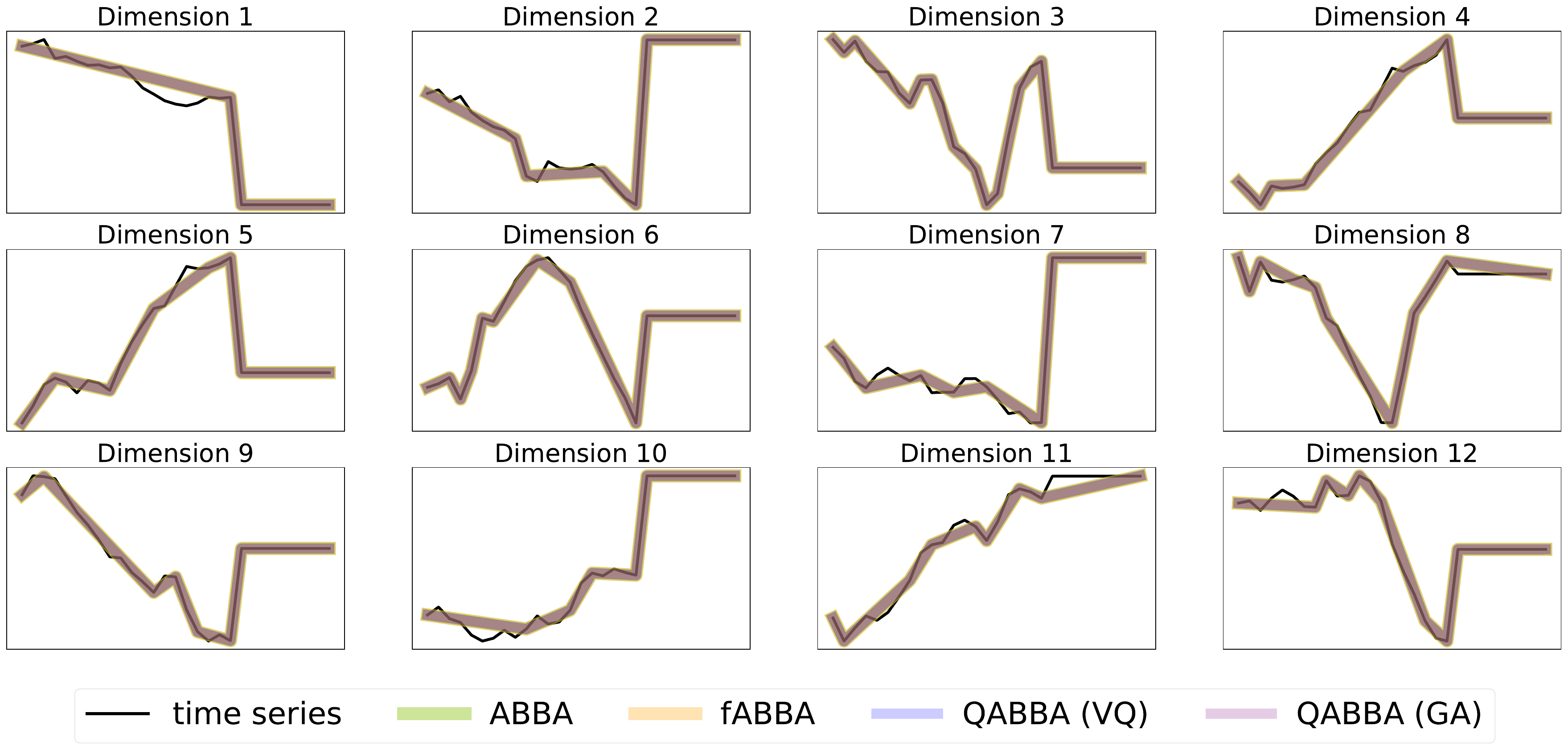}}\\[0.5em]
\subfloat[NATOPS]{\includegraphics[width=0.55\textwidth]{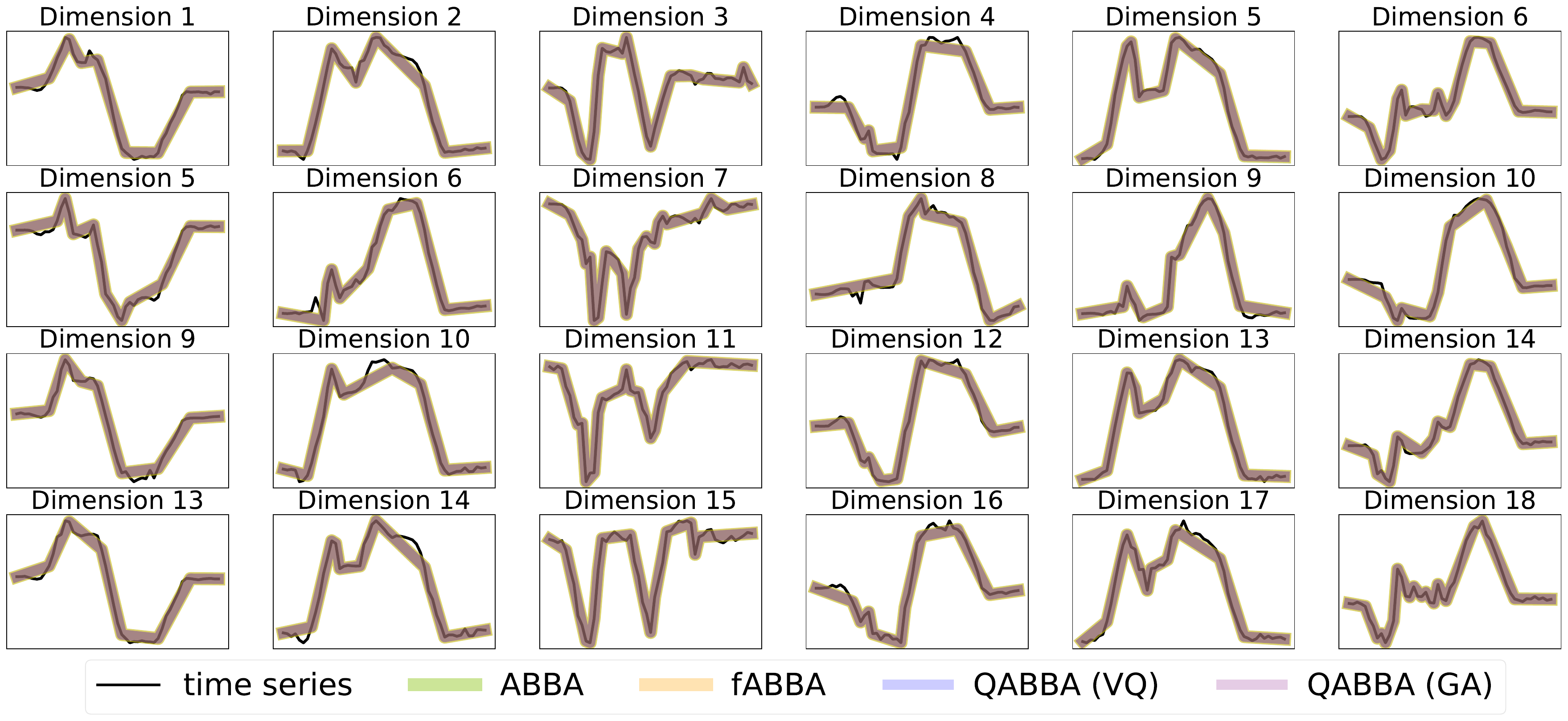}}
\caption{Symbolic reconstruction of UEA Archive (II).}
\label{fig:reconstUEA2}
\end{figure}

\rev{The UEA Archive contains a wide range of multivariate time series from real-world applications. Directly employing ABBA, fABBA, and QABBA (GA \& VQ) on these datasets is not possible, since the consistency of symbols is not guaranteed. We use the joint symbolic representation introduced in \cite{chen2024joint} to enable the computation of symbolic approximations of multivariate time-series data in a unified manner. The approximation error is measured by the distance between ground-truth data and reconstructed data, computed as the average squared difference over the multidimensional tensor. All algorithms use the same number of symbolic centers. We use the storage ratio defined in Section~\ref{sec:store} to evaluate the storage savings. Experiments using two compression settings ($\tol = 0.005$ and $\tol = 0.05$) are shown in \figurename~\ref{fig:UEA1} and \figurename~\ref{fig:UEA4}.}

\rev{\figurename~\ref{fig:UEA4} reports the storage ratio and its component contributions, showing that the storage savings of QABBA are significant: the required storage is nearly half of that required without quantization, at the cost of only a negligible increase in reconstruction error. In addition, \figurename~\ref{fig:UEA4} shows that a large portion of the storage of these methods is devoted to strings; therefore, further compression can be obtained from Lempel-Ziv-Welch (LZW) \cite{1659158}.}

\begin{table}[ht]
\centering
\caption{Selected datasets from the UEA Archive.}
\label{tab:UEAinfo}
\scriptsize
\setlength{\tabcolsep}{6pt} 
\setlength{\extrarowheight}{2pt} 
\begin{tabular}{l r r r} 
\toprule
Dataset & Size & Dimension & Length \\
\midrule
AtrialFibrillation & 30 & 2 & 640 \\
BasicMotions & 80 & 6 & 100 \\
CharacterTrajectories & 2,858 & 3 & 182 \\
Epilepsy & 275 & 3 & 206 \\
JapaneseVowels & 640 & 12 & 29 \\
NATOPS & 360 & 24 & 51 \\
UWaveGestureLibrary & 440 & 3 & 315 \\
\bottomrule
\end{tabular}
\end{table}

\rev{The approximation error of QABBA compared to the non-quantized variants (e.g., comparing ABBA with QABBA (VQ), and fABBA with QABBA (GA)) remains similar in terms of DTW distance but can be significantly higher in terms of MSE. We can thus conclude that quantization causes more error as measured by MSE than as measured by DTW. We note that our bit-widths for the length and increment values were taken from our experiments with synthetic data; it is likely that, in practice, these parameters should be tuned to a specific dataset, which can improve reconstruction error.}

\rev{Acceptable error can be evaluated perceptually by plotting both the raw time series and the reconstruction; if they are close enough, this may be considered ``acceptable''. Both ABBA and fABBA are well-established methods for high-quality symbolic reconstruction, and our intent is to validate the reconstruction error by comparing QABBA with ABBA or fABBA. Hence, following the settings described in Section~\ref{sec:uea}, we plot the reconstructions of ABBA, fABBA, QABBA (VQ), and QABBA (GA) with $\tol=0.05$ for the first multivariate time series of each UEA dataset in \tablename~\ref{tab:UEAinfo}. The reconstruction comparisons are shown in \figurename~\ref{fig:reconstUEA1} and \ref{fig:reconstUEA2}. The reconstructions of ABBA and QABBA almost overlap with the original time series, which demonstrates that all tested methods achieve satisfactory reconstruction and that the comparison of their storage is justified.}

}

\begin{figure}[ht!]
\centering
\subfloat[$\tol = 0.005$.]{\includegraphics[width=0.27\textwidth]{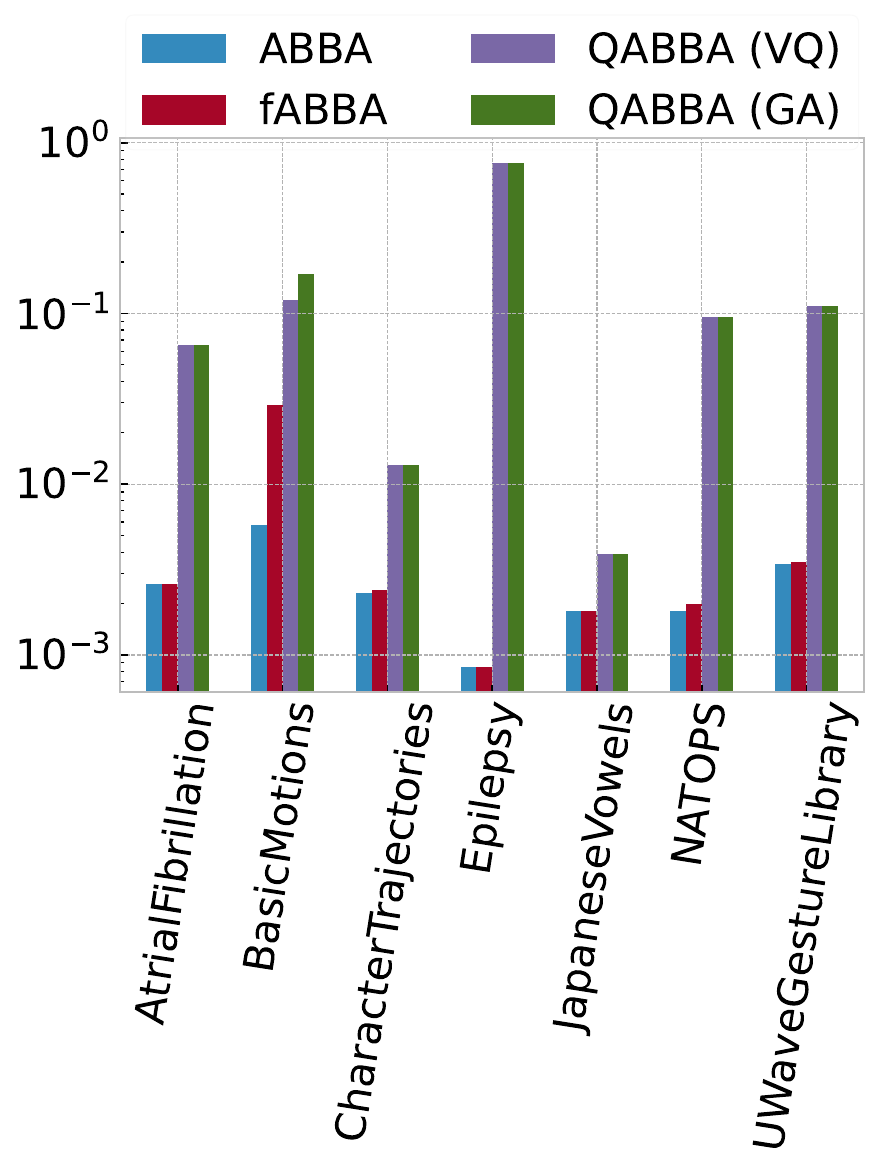}}
\subfloat[$\tol = 0.05$.]{\includegraphics[width=0.27\textwidth]{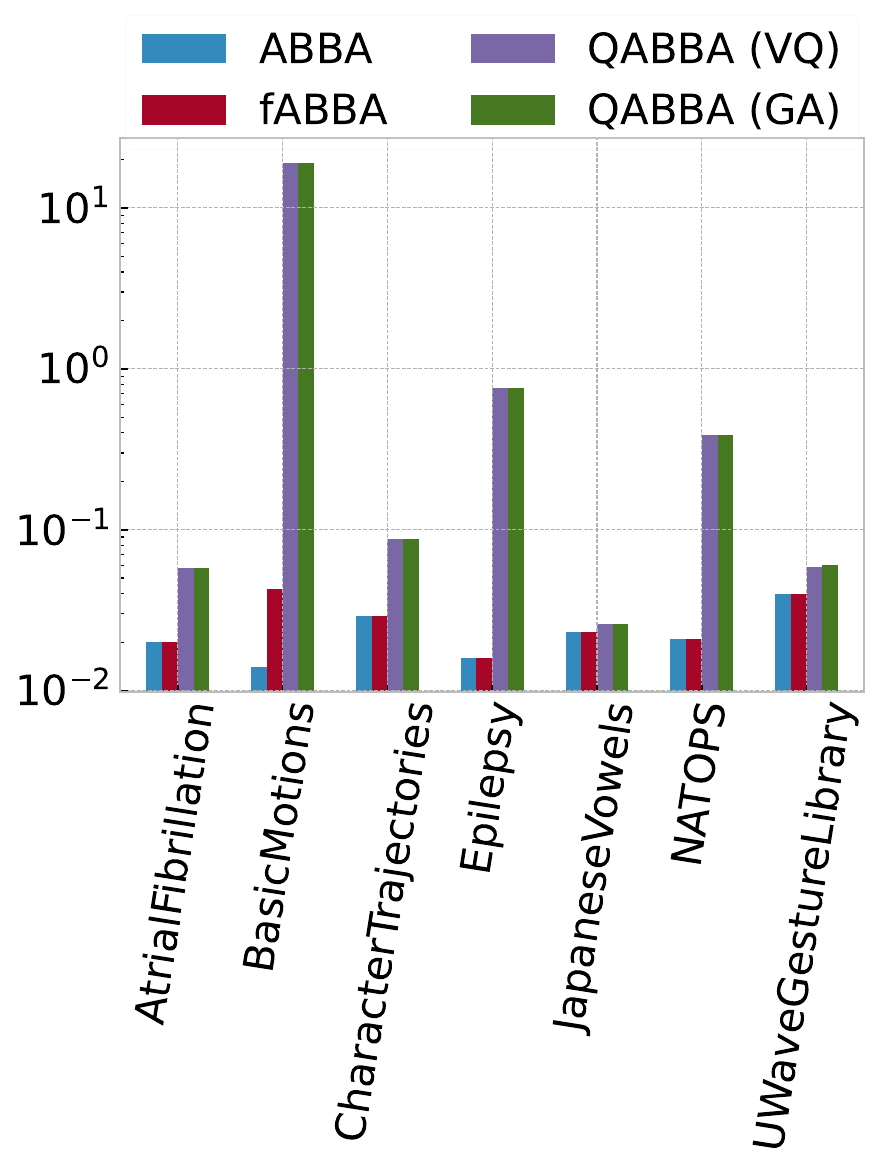}}
\caption{\text{Errors of reconstruction}.}
\label{fig:UEA1}
\end{figure}

\begin{figure}[ht!]
	\centering
	\subfloat[$\tol = 0.005$]{\includegraphics[width=0.29\textwidth]{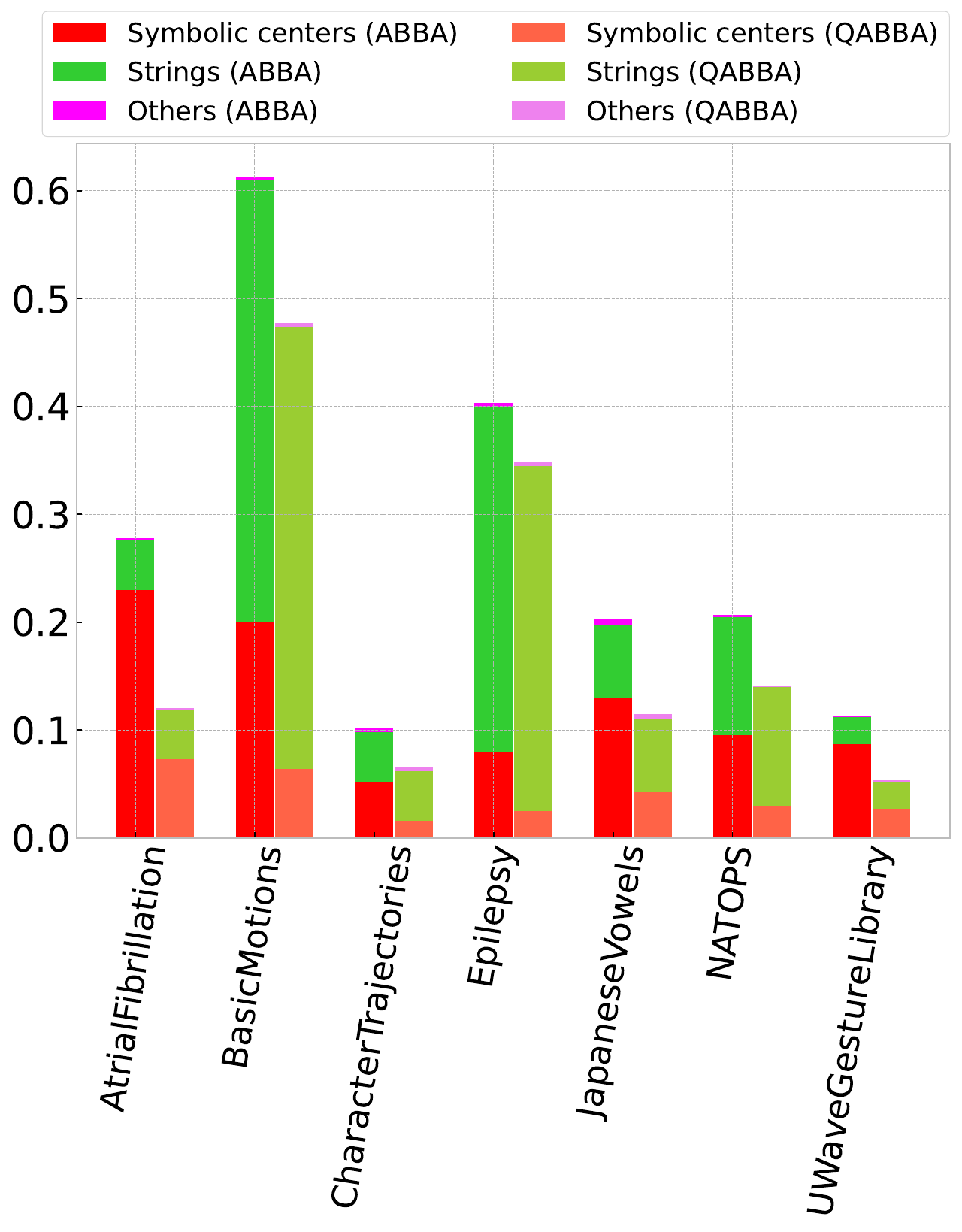}}
	\subfloat[$\tol = 0.05$]{\includegraphics[width=0.29\textwidth]{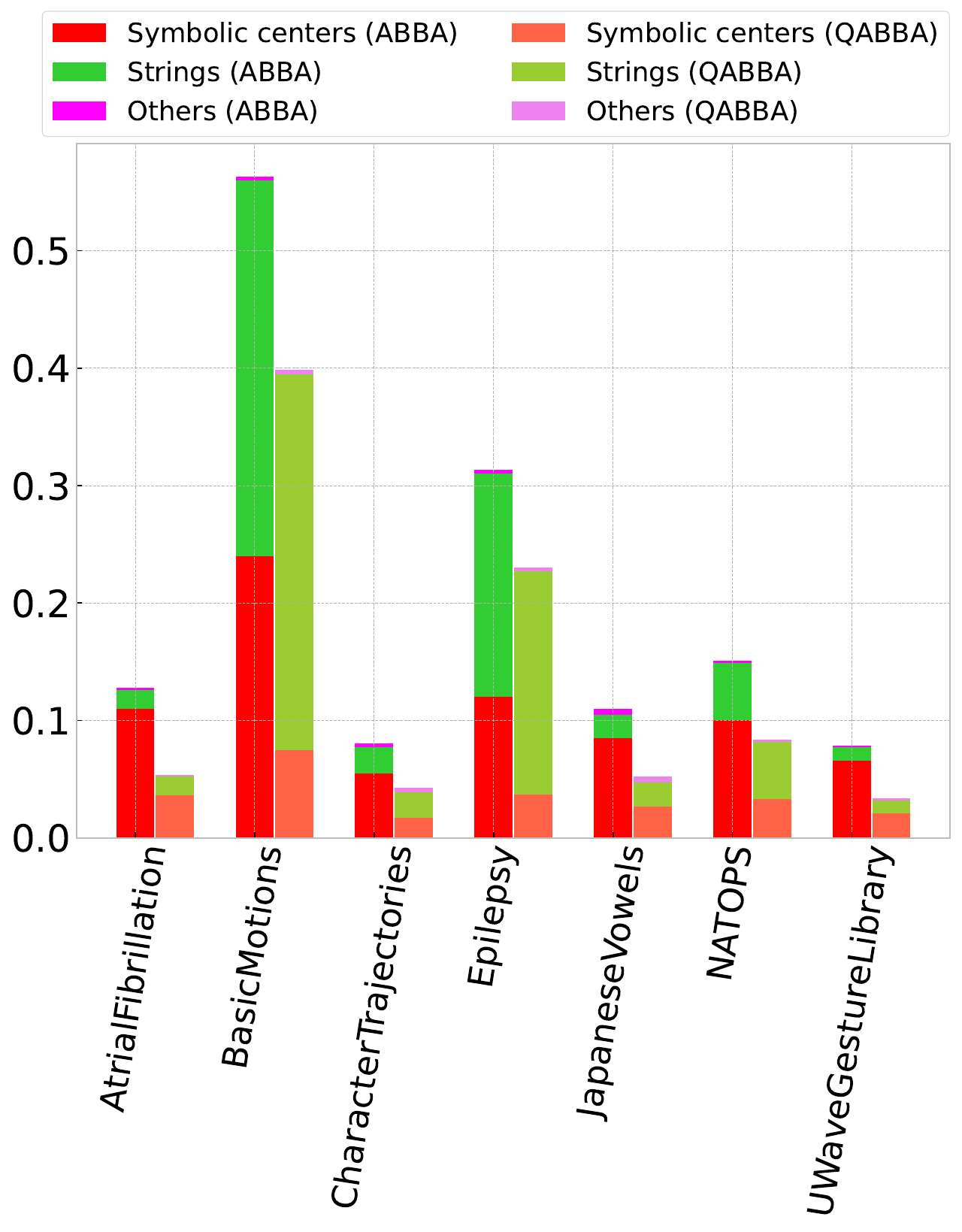}}
		\caption{\rev{Storage ratio.}}

\label{fig:UEA4}
\end{figure}

\subsection{\rev{Storage-Oriented Compression Analysis}}\label{sec:compression_exp}

\rev{The storage analysis in Section~\ref{sec:store} assumes that the symbolic sequence is stored using a byte-addressed representation. This is a conservative and implementation-friendly convention, but it can be wasteful when the alphabet size is substantially smaller than $2^8$ or when repeated symbolic patterns appear in the sequence. We therefore conduct an additional compression study to answer two questions: (i) whether a lossless second-stage codec can further reduce the QABBA representation at the symbol level, and (ii) where QABBA lies in the rate--distortion landscape relative to representative lossy time-series compressors.}

\rev{For the symbol-level study, the fitted QABBA representation is kept fixed and only the symbolic sequence is recoded. We compare the byte-symbol baseline with fixed-width symbol packing, Huffman coding, Lempel-Ziv-Welch (LZW), and common entropy or dictionary codecs. Since these codecs operate after QABBA has produced its symbolic approximation, they are lossless with respect to the symbolic sequence and do not change the reconstructed time series. For the lossy-compression comparison, we evaluate QABBA together with piecewise aggregate approximation (PAA), discrete-Fourier-transform (DFT) low-pass approximation, piecewise linear approximation (PLA), and open-source scientific data compressors including SZ3~\cite{Zhao2021SZ3}, ZFP~\cite{Lindstrom2014ZFP}, and Serf~\cite{Li2025Serf}. Storage ratio is measured as compressed bits divided by the raw float32 storage.}

\begin{figure}[htb!]
	\centering
	\subfloat[\rev{Rate--distortion comparison with lossy baselines.}]{\includegraphics[width=0.78\textwidth]{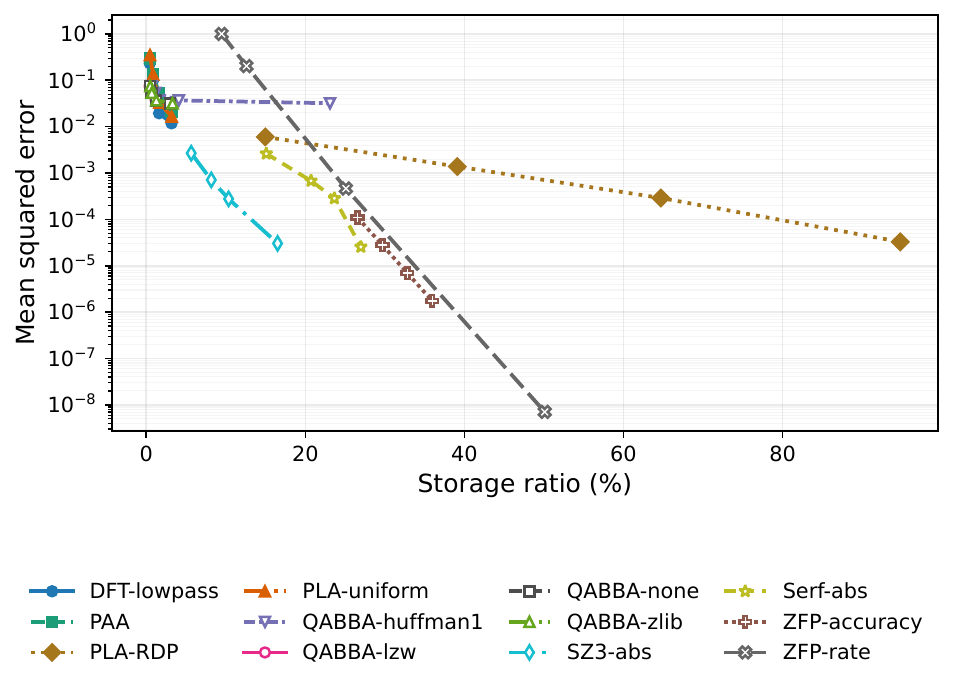}}\\[0.75em]
	\subfloat[\rev{Symbol-level lossless codec comparison.}]{\includegraphics[width=0.78\textwidth]{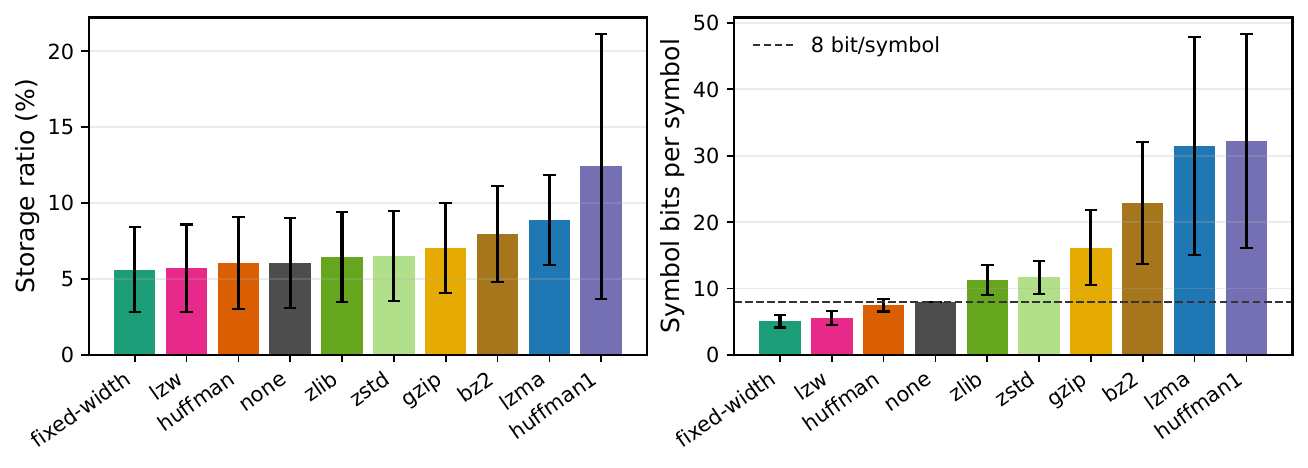}}
		\caption{\rev{Compression behavior of QABBA under second-stage symbol coding and against representative lossy time-series compressors. In (a), lower-left points indicate better rate--distortion tradeoffs. In (b), the fixed-width codec denotes alphabet-aware bit packing using $\lceil \log_2 |\mathcal{L}| \rceil$ bits per symbol, rather than a statistical compressor.}}
\label{fig:comp_new}
\end{figure}

\rev{\figurename~\ref{fig:comp_new}(b) shows that the byte-symbol convention leaves measurable redundancy in the symbolic sequence. Fixed-width packing and LZW are the most effective second-stage choices in this setting. Across the tested series, fixed-width packing reduces the average storage ratio from $0.0607$ for uncompressed byte symbols to $0.0561$, while LZW achieves $0.0572$; equivalently, the median symbol cost decreases from $8$ bits/symbol to about $5.5$ and $6.05$ bits/symbol, respectively. The improvement is obtained without changing the approximation error, because only the representation of the already-produced symbol string is altered. By contrast, general-purpose container codecs such as gzip, bzip2, and lzma can be less effective on short QABBA strings because their headers, dictionaries, and model overheads may dominate the payload. These results suggest that fixed-width packing is a strong default when the alphabet size is known, whereas LZW is attractive when repeated motifs create additional dictionary structure.}

\rev{\figurename~\ref{fig:comp_new}(a) clarifies the role of QABBA as a very-low-rate lossy representation. With LZW symbol coding, QABBA attains a median storage ratio of $0.0060$ and a median compression factor of approximately $168\times$, while preserving the same median MSE as the uncompressed-symbol QABBA representation ($4.5\times 10^{-2}$). Classical dimensionality-reduction baselines such as PAA, DFT low-pass approximation, and uniform PLA occupy a nearby but higher-rate regime, with median storage ratios around $0.012$ and median compression factors around $89$--$91\times$. Scientific floating-point compressors show a different operating point: SZ3, Serf, and ZFP can achieve substantially lower reconstruction errors under accuracy-oriented settings, but their median storage ratios are much higher in this experiment, ranging from about $0.095$ to $0.311$. Thus, QABBA should not be viewed as a replacement for error-bounded floating-point compressors when extremely small numerical error is required. Rather, it provides an orthogonal symbolic-compression regime: it aggressively reduces storage, preserves interpretable symbolic structure for downstream mining tasks, and can be strengthened by a lossless second-stage codec without modifying the QABBA approximation pipeline.}

{
\section{Disadvantages}\label{sec:short}
\rev{QABBA’s symbolic representation does not fully leverage the extensive array of discrete operations utilized by methods like SAX, for example, i) grammar inference over SAX words \cite{zhang2020semantic}; ii) graph construction on time series \cite{Gao2019}; iii) regular expression matching on time series \cite{10.2312:eurova.20221081}; iv) time-series hashing \cite{10.1145/956750.956808}; and v) time-series mapping to suffix trees \cite{lin2004visually}. We acknowledge that QABBA, together with its predecessor ABBA, is not designed to support the complete range of these discrete operations, which are optimized for SAX’s fixed-resolution discretization approach. However, because QABBA’s symbolic sequence is structurally identical to SAX’s output, we respectfully assert that QABBA’s symbolic representation retains considerable value for downstream data-mining applications.}

}
\section{Summary and future work}\label{sec:final}
\rev{In this paper, we present QABBA, a quantization-based symbolic approximation method for time series, and explore its performance in terms of reconstruction error and storage efficiency on various datasets. We theoretically and empirically validate the possibility and potential of using quantization in ABBA. We investigate the bit-width required for quantization to ensure acceptable symbolic reconstruction and provide a theoretical understanding of the relationship between quantization error and symbolic-approximation error; this analysis can also be applicable to other quantization algorithms. The empirical results demonstrate that QABBA requires only a fraction of the storage needed for symbolic approximation while maintaining acceptable accuracy compared with ABBA and fABBA, with no observed speed loss in our Python-level runtime profiles. Our focus is to preserve the core time-series structure and patterns under additional compression, which can be further enhanced with LZW compression. We believe that lower-bitwidth representations may be promising in certain scenarios. We also investigate time-series regression with QABBA and language models, demonstrating the potential benefits of our symbolic approximation technique. Semantic understanding of time series using QABBA is promising for many time-series mining tasks, and further study will be left as future work.}

\section*{Acknowledgements}\phantomsection
The first author acknowledges funding from the European Union (ERC, inEXASCALE, 101075632). Views and opinions expressed are those of the authors only and do not necessarily reflect those of the European Union or the European Research Council. Neither the European Union nor the granting authority can be held responsible for them. Additionally, the first author acknowledges funding from the Charles University Research Centre program (No.\ UNCE/24/SCI/005). The second author acknowledges funding from the France 2030 NumPEx Exa-MA project (ANR-22-EXNU-0002), managed by the French National Research Agency (ANR). {The last author acknowledges funding from the Research Center Informatics (No.
CZ.02.1.01/0.0/0.0/16 019/0000765) and Brain Dynamics (No.
CZ.02.01.01/00/22 008/0004643)}.

\section*{Declarations}


\begin{itemize}
\item Funding: The first author was funded by the Charles University Research Centre program No. UNCE/\-24/\-SCI/005. The first and second authors were funded by the European Union (ERC, inEXASCALE, 101075632). Views
and opinions expressed are those of the authors only and do not necessarily reflect those of the European Union or the European Research Council. Neither the European Union nor the granting authority can be held responsible for them. {We acknowledge the experimental data from \url{https://www.timeseriesclassification.com/}}.
\item Competing interests: The authors have no competing interests to declare that are relevant to the content of this article.
\end{itemize}

\bibliographystyle{unsrtnat}

\bibliography{cas-refs}

\end{document}